\documentclass[11pt]{article}

\usepackage[margin=1in]{geometry}
\usepackage{amsmath,amssymb,amsfonts}
\usepackage{graphicx}
\usepackage{booktabs}
\usepackage{natbib}
\usepackage{hyperref}
\usepackage{xurl}   
\usepackage{algorithm}
\usepackage{algpseudocode}
\usepackage{multirow}
\usepackage{caption}
\usepackage{subcaption}
\usepackage{placeins}

\graphicspath{{sections/figures/}}

\newcommand{\vxi}{\boldsymbol{\xi}}

\newcommand{\mhat}[1]{\hat{\mathbf{#1}}}
\newcommand{\R}{\mathbb{R}}
\newcommand{\norm}[1]{\left\lVert #1 \right\rVert}

\newcommand{\maineqref}[1]{\eqref{#1}}

\title{Validated Synthetic Patient Generation for Small Longitudinal Cohorts: \\
Coagulation Dynamics Across Pregnancy}

\author{
Jeffrey D.\ Varner$^{1,*}$,
Maria Cristina Bravo$^{2}$,
Carole McBride$^{3}$,
\\
Thomas Orfeo$^{2}$,
Ira Bernstein$^{3}$
\\[6pt]
{\small $^{1}$Robert F.\ Smith School of Chemical and Biomolecular Engineering,} \\
{\small Cornell University, Ithaca, NY 14850} \\
{\small $^{2}$Department of Biochemistry, The Robert Larner, M.D.} \\
{\small College of Medicine, University of Vermont Medical Center, Burlington, VT 05401} \\
{\small $^{3}$Department of Obstetrics, Gynecology, and Reproductive Sciences,} \\
{\small The Robert Larner, M.D.\ College of Medicine,} \\
{\small University of Vermont Medical Center, Burlington, VT 05401} \\[6pt]
{\small $^{*}$Corresponding author: jdv27@cornell.edu}
}

\date{}

\begin{document}
\maketitle

\begin{abstract}
Small longitudinal cohorts, common in maternal health, rare diseases, and early-phase trials,
limit computational modeling because enrollment is slow and the data are too sparse to train
reliable models. We present multiplicity-weighted Stochastic Attention (SA), a generative
framework based on modern Hopfield networks. Stochastic attention stores real patient profiles
as memory patterns in a continuous energy landscape. The resulting distribution is a
finite mixture with one component centered on each stored profile. We generated the reported
cohort with Langevin dynamics; direct sampling from the same distribution reproduced its
fidelity and membership-inference results. Multiplicity weights amplify selected subgroups at
inference time without retraining. We applied the method to longitudinal coagulation data from
23 patients, with 72 features measured before pregnancy and during the first and third
trimesters. The cohort included three patients with polycystic ovary syndrome and five who
developed preeclampsia. Across statistical, structural, and mechanistic tests, including
a separately specified coagulation model that was blind to the generator but calibrated
on the same cohort, the synthetic profiles closely matched the real profiles under the
measures applied at this sample size. A model calibrated on synthetic profiles predicted
held-out real-patient TGA measurements as well as one calibrated on real profiles. These
results provide proof of concept for using SA in the tested modeling tasks, including
mechanistic calibration and hypothesis generation, with very small longitudinal cohorts.

\medskip
\noindent\textbf{Keywords:} synthetic data generation, stochastic attention, modern Hopfield networks,
multiplicity weighting, coagulation, pregnancy, longitudinal data, mechanistic validation

\end{abstract}

\section{Introduction}
\label{sec:intro}

In 2021, the United States maternal mortality rate reached 32.9 deaths per 100,000
live births, the highest level since 1965~\citep{Toy2023WSJ,Harris2023}. Between 2018 and 2022,
Non-Hispanic Black women died at 2.8 times the rate of White women, and
disorders related to pregnancy, including hemorrhage and venous
complications, were the leading underlying cause of death, accounting for
17.4\% of 6{,}283 pregnancy-related deaths~\citep{Chen2025}. These figures
highlight the role of the coagulation system in maternal outcomes. Blood coagulation is a complex enzymatic cascade in which a small
tissue factor (TF) stimulus triggers a series of serine protease activation reactions
that ultimately produce thrombin, the central effector of
hemostasis~\citep{Mann2011,Butenas2009}. Thrombin cleaves fibrinogen into fibrin,
activates platelets, and amplifies its own production through positive feedback via
Factors~V, VIII, and XI, while simultaneously initiating its own shutdown through the
thrombomodulin--protein~C anticoagulant pathway~\citep{Hockin2002,Bravo2012}. Pregnancy
alters this hemostatic balance, shifting toward a hypercoagulable state
that prepares for delivery~\citep{Hellgren2003,Grouzi2022,Bremme2003}.
Procoagulant factors including fibrinogen, Factor~VIII, and von Willebrand factor
increase substantially across gestation, while natural anticoagulants such as
antithrombin and protein~S decline. Two conditions further shift this balance in ways that are not fully understood.
Polycystic ovary syndrome (PCOS), a hormonal and metabolic disorder affecting
approximately 6.6\% of reproductive-age women~\citep{Azziz2004}, has been associated with
impaired fibrinolysis and a prothrombotic tendency~\citep{Palomba2015}. Preeclampsia (PE), a hypertensive
disorder that complicates 2--8\% of pregnancies worldwide~\citep{Abalos2014}, is
characterized by endothelial dysfunction, platelet activation, and altered thrombin
generation~\citep{Dusse2011}. Mathematical models of the coagulation cascade have provided mechanistic insight
into patient-specific thrombin generation~\citep{Luan2007,Luan2010}. These range
from the original Hockin--Mann stoichiometric framework~\citep{Hockin2002} through
extensions incorporating the thrombomodulin--protein~C pathway~\citep{Bravo2012}
to the Brummel-Ziedins 2012 (BZ2012) model with detailed activated protein~C
(APC)-mediated Factor~Va inactivation
kinetics~\citep{BrummelZiedins2012}. However, applying these models to pregnancy
cohorts requires longitudinal datasets capturing coagulation factors, inhibitors,
and thrombin generation assay (TGA) measurements across gestation, data that is
rare and expensive to collect.

The core barrier is sample size. Longitudinal pregnancy coagulation studies
typically enroll tens of patients with complete follow-up across multiple trimesters,
yielding datasets where the number of features~$p$ (coagulation factors, TGA
parameters, viscoelastic measurements) exceeds the number of patients~$n$. Rare
complications compound this problem. Small longitudinal cohorts inevitably contain
only a handful of patients with any given condition, far too few for
condition-specific statistical analysis. This $n < p$ regime limits conventional
approaches because covariance matrices are rank-deficient, cross-validation is
unreliable, and overfitting is nearly inevitable~\citep{Hastie2009}. Synthetic data can
provide dense simulated inputs for mechanistic modeling, workflow testing, and hypothesis
generation~\citep{Gonzales2023,Kuhnel2024}. A further motivation was to support the
development of deep learning models. However, existing methods face
limitations in the small-cohort longitudinal setting. Sampling from a fitted
multivariate normal (MVN) distribution is singular when $n < p$ and must be
regularized~\citep{Ledoit2004}, introducing bias that distorts the joint distribution.
The multivariate-normal model also assumes Gaussian marginals and linear correlations and
cannot amplify a specific subpopulation while preserving
its signatures. More expressive models such as generative adversarial networks (GANs) and
variational autoencoders (VAEs)~\citep{Goodfellow2014,Kingma2014,Xu2019CTGAN}, including recent longitudinal
extensions~\citep{Kuhnel2024}, require substantially larger training sets and are
prone to mode collapse at small $n$~\citep{DeMitri2025} (we confirm this empirically for Conditional Tabular GAN (CTGAN) at
$n{=}23$ in this work). A generative method is needed that can operate
directly on the geometry of a small dataset without estimating a full
covariance model.

Stochastic Attention (SA) is such a framework. Rather than fitting a full
covariance model or training a neural generator, SA treats the stored patient
profiles themselves as the model. Building on modern Hopfield network
theory~\citep{Ramsauer2021}, each profile becomes a memory pattern in a
continuous energy landscape. This energy induces a continuous probability
distribution that is exactly a finite Gaussian mixture in the reduced space,
with one nonzero-variance component centered on each stored pattern. Sampling
from this distribution therefore produces continuous variations around the
stored patterns rather than exact copies. Because sampling occurs in a
principal component analysis (PCA)-reduced linear subspace anchored on the observed cohort, SA never forms a
full $p \times p$ covariance matrix and so avoids both the rank deficiency that
limits MVN and the mode collapse that limits GAN/VAE approaches when $n < p$.
We recently introduced a multiplicity-weighted extension of the Hopfield energy
that attaches a per-pattern weight to each stored profile~\citep{Varner2026Multiplicity},
which allows continuous interpolation
between unconditioned generation and targeted amplification of a rare
subpopulation at inference time, without retraining. This study asked whether SA could
faithfully generate \emph{longitudinal}
trajectories from a small cohort. In earlier proceedings work, we paired our
mechanistic coagulation pipeline with per-visit MVN sampling to produce
synthetic populations~\citep{Varner2023JuliaCon}; that approach captured
single-visit marginals but, by fitting each visit independently, generated
patients whose trajectories across gestation were statistically incoherent. Small
longitudinal cohorts require a generator that preserves this joint structure.

In this study, we used multiplicity-weighted SA to generate complete
longitudinal patient profiles from a small pregnancy coagulation cohort
($K{=}23$ patients, 72 features per visit, 3 visits) that includes rare
clinical subgroups (PCOS, $n{=}3$; Developed~PE, $n{=}5$). Our pipeline
concatenates each patient's multi-visit profile into a single vector, applies
PCA for dimensionality reduction, and generates new patients via
multiplicity-weighted Langevin dynamics on the Hopfield energy landscape, with
a direction-magnitude decomposition that preserves the natural variance
structure of continuous clinical measurements. We then asked whether the
synthetic patients reproduced marginal feature distributions, the cross-visit
covariance structure, condition-specific signatures of rare clinical
subgroups, and the biological consistency obtained when the profiles were processed by
a separately specified, generator-blind ordinary differential equation (ODE) model
of the coagulation cascade that was calibrated on the same real
cohort~\citep{BrummelZiedins2012}. As a downstream-utility test, we
asked whether a mechanistic model calibrated entirely on synthetic patients
could predict held-out real-patient TGA measurements. As a proof of concept, the
results show that SA recovered the statistical structure of this cohort and matched its
mechanistic plausibility under the tests used here. Generalization to other clinical cohorts
remains to be tested.

\section{Results}
\label{sec:results}

We generated $N{=}100$ synthetic longitudinal patient profiles using SA and compared
them against the $K{=}23$ real patients across four validation levels. As a baseline, we generated
$N{=}100$ patients from a regularized multivariate normal (MVN) distribution fitted
to the same data.

\subsection{Marginal Plausibility}
\label{sec:results-l1}

We first assessed whether SA-generated patients reproduced per-feature summary
statistics across all three visits. The pooled median mean relative error
(MRE) across all 216 feature--visit entries was 1.2\% (95\% bootstrap CI:
1.0--1.6\%), with 193 of 216 entries (89\%) below 5\%
(Table~\ref{stab:mre-summary}), indicating that SA captured the central
tendencies of the real cohort. Per-visit MREs for representative
coagulation features are summarized in Table~\ref{tab:summary}, where
Factor~II and Factor~VIII fell below 2\%, antithrombin below 3\%, and
fibrinogen below 1\% across all three visits. Marginal agreement alone could result from
profiles that closely copied stored records, so we also quantified similarity to stored
profiles with two metrics (Table~\ref{stab:memorization}). The mean novelty score
($1 - \max_k \cos(\hat{\vxi}, \hat{\mathbf{m}}_k)$) was 0.50 at
$\beta = \beta^*$. The median
nearest-neighbor distance from each synthetic patient to its closest real
patient was 6.14 standardized units, compared with 6.87 for the median
real-to-real nearest-neighbor distance, a ratio of 0.89 indicating that
synthetic patients were approximately as far from real patients as real
patients were from each other. Thus, stochastic attention matched the central tendencies
without reproducing stored profiles exactly.

\begin{figure}[p]
\centering
\includegraphics[width=\textwidth]{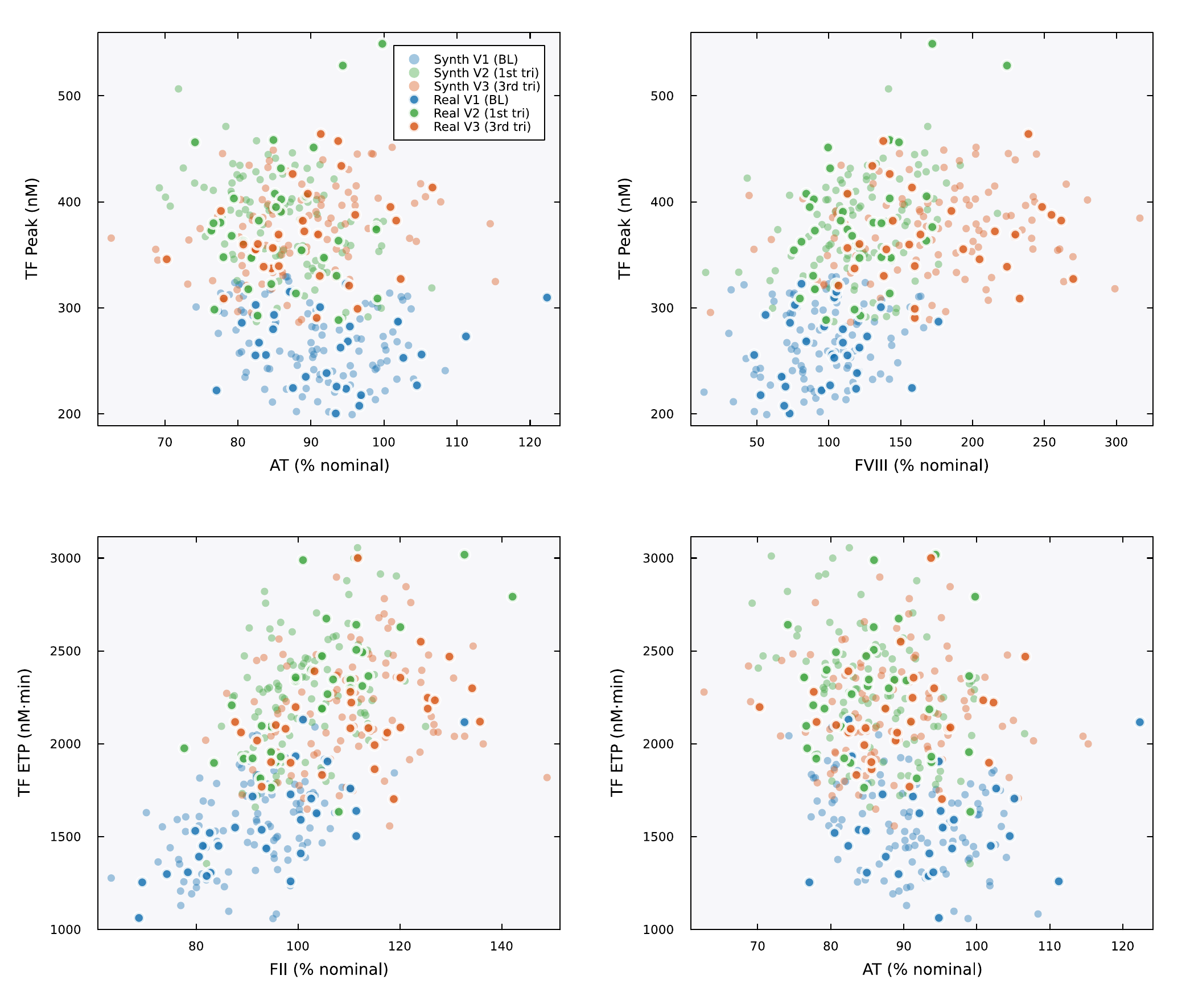}
\caption{\textbf{Physiological correlations.}
Scatter plots of coagulation factor levels versus thrombin generation parameters
across all three visits. Filled markers are real patients; open markers are
SA-generated synthetic patients. Colors encode visit: blue (V1/baseline),
green (V2/first trimester), orange (V3/third trimester). Real and synthetic
patients occupy the same joint regions across all four pairwise relationships. AT
inversely correlates with TF-initiated peak thrombin (top left), FVIII
positively correlates with peak (top right), and both FII (bottom left) and AT
(bottom right) show the expected relationships with ETP. The visit-dependent
shift in these relationships, reflecting the pregnancy-driven increase in
procoagulant factors, is also preserved in the synthetic cohort.}
\label{fig:bio-correlations}
\end{figure}

\begin{figure}[p]
\centering
\includegraphics[width=\textwidth]{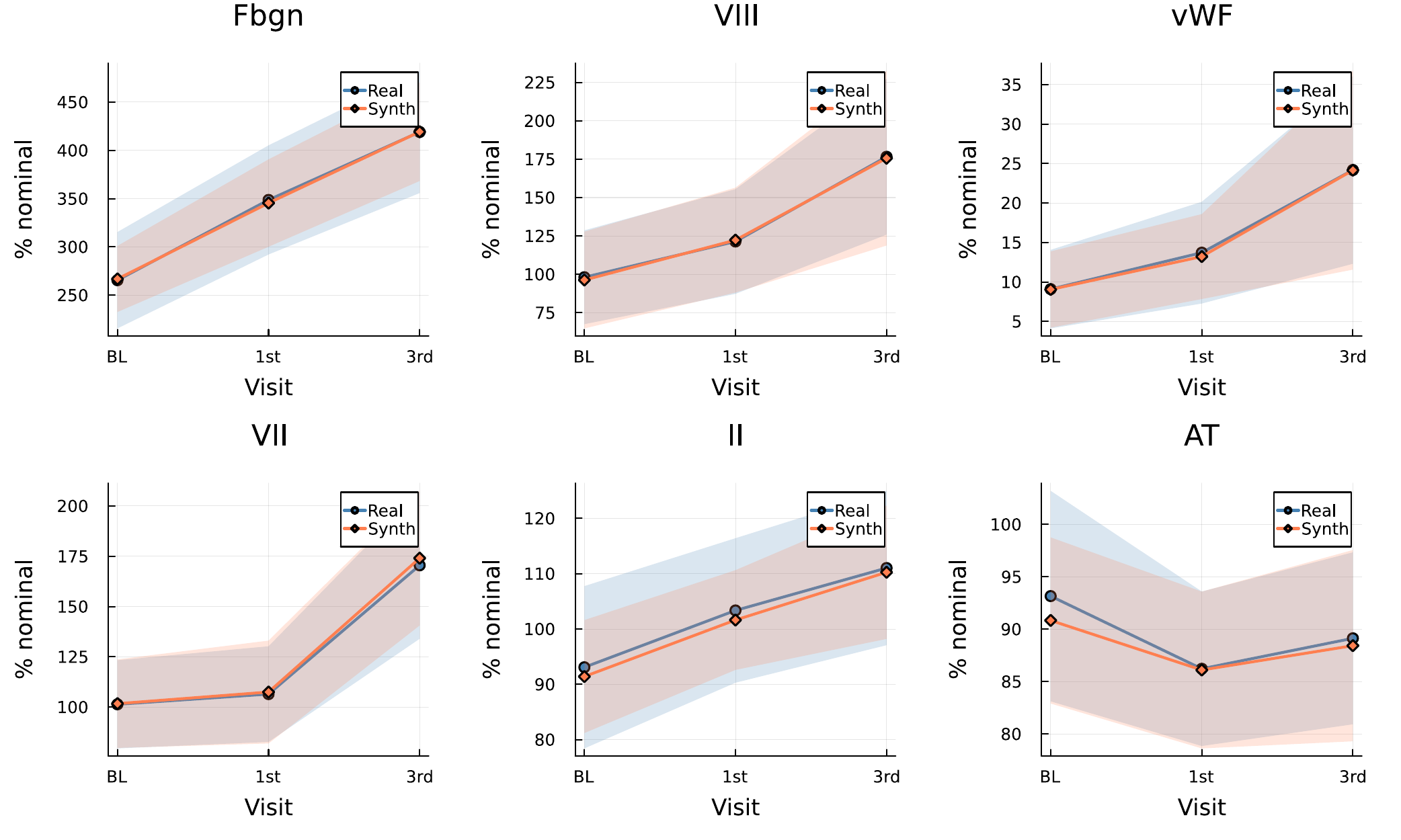}
\caption{\textbf{Pregnancy-related longitudinal trajectories.}
Mean $\pm$ standard deviation bands for six key coagulation features across
visits (BL = baseline, 1st = first trimester, 3rd = third trimester) for real
(blue) and SA-generated synthetic (orange) patients. The characteristic
pregnancy-driven increases in fibrinogen, Factor~VIII, and vWF are captured, as
are the stable-to-declining patterns in Factor~II and AT.}
\label{fig:pregnancy-progression}
\end{figure}

We next examined whether known physiological relationships and longitudinal
trajectories were preserved in the synthetic cohort. Pairwise scatter plots of
coagulation factor levels against thrombin generation parameters showed that
SA-generated patients occupied the same joint regions as real patients
(Fig.~\ref{fig:bio-correlations}). The expected inverse relationship between
antithrombin and TF-initiated peak thrombin was preserved, as were the positive
associations between Factor~VIII and peak, prothrombin and endogenous thrombin
potential (ETP), and the inverse relationship between antithrombin and ETP.
Per-visit means and standard deviations for six key coagulation features showed
that synthetic patients tracked the real longitudinal trajectories closely, with
overlapping confidence bands at all three visits
(Fig.~\ref{fig:pregnancy-progression}). Fibrinogen, Factor~VIII, and von Willebrand
factor showed the expected monotonic increases from baseline through the third
trimester in both cohorts, while Factor~II increased modestly and antithrombin
declined then partially recovered, consistent with the known hypercoagulable shift
of pregnancy. The 72 features also included viscoelastic (ROTEM) and fibrinolytic
parameters, which had comparable MREs to the coagulation factors
(Table~\ref{stab:mre-summary}). These variables influenced the generated profiles
through their loadings on the retained principal components. Together, these results
showed that the synthetic cohort preserved the tested physiological relationships and
the direction and magnitude of longitudinal change.

To determine whether standard deep tabular models provided a viable baseline at this sample
size, we compared SA against Conditional Tabular GAN (CTGAN) and Tabular Variational Autoencoder
(TVAE)~\citep{Xu2019CTGAN} (Table~\ref{stab:ctgan}). The CTGAN model produced median
MREs of approximately 19\% across all tested epoch counts (300--3,000), consistent with
mode collapse at this sample size. The TVAE model achieved comparable marginal fidelity at
high epoch counts (median MRE
1.8\% at 3,000 epochs), but operated on per-visit rows ($n{=}90$) rather than
concatenated longitudinal profiles ($n{=}23$), meaning it could not capture cross-visit
covariance structure and had no mechanism for conditional subgroup generation. Thus, neither
deep model provided a suitable complete-profile baseline for this $23$-patient cohort.

To examine the role of the shared PCA representation, we compared stochastic attention
with a fitted two-component diagonal-covariance Gaussian mixture, a kernel-density estimator, $k$-nearest-neighbor interpolation, and a
Gaussian copula in the same $d{=}18$ PCA subspace
(Supplementary Methods, PCA-space ablation; Supplementary Table~\ref{stab:pca-ablation}).
Nearest-neighbor samples remained close to
stored profiles: median novelty, $1-\max_k\cos(\hat{\vxi},\hat{\mathbf{m}}_k)$, was $0.05$,
and only $5\%$ of the synthetic patients exceeded the $0.2$ novelty cutoff.
The kernel-density estimator had marginal error $1.9\%$ and mechanistic overlap $0.82$ (the fraction of pooled synthetic
predicted-to-recorded TGA ratios within the real 5th--95th percentile range). The two-component mixture and
copula matched stochastic attention more closely in marginal error ($1.1\%$ and $1.3\%$
versus $1.2\%$) and mechanistic overlap ($0.92$ and $0.93$ versus $0.90$). All profiles from
stochastic attention and the copula exceeded the novelty cutoff, compared with $94\%$ from
the two-component mixture. Stochastic attention had a larger empirical cross-visit correlation error
than the two-component mixture and copula ($0.64$ versus $0.42$ for both). This metric measured
agreement with the correlation matrix of the same $23$ patients used to construct the
stochastic-attention memory matrix or fit the baseline generators; it therefore assessed
in-sample reproduction rather than recovery of the unknown population correlation matrix.
Holding the PCA representation constant showed that the marginal errors and mechanistic
overlaps achieved by the two-component mixture and copula were similar to those of stochastic
attention and therefore were not specific to the stochastic-attention generator. In contrast,
empirical correlation error and novelty differed across generators. Thus, generator choice
affected these aspects of performance, and no generator performed best on every measure.

\subsection{Cross-Visit Covariance Structure}
\label{sec:results-l2}

\begin{figure}[p]
\centering
\includegraphics[width=\textwidth]{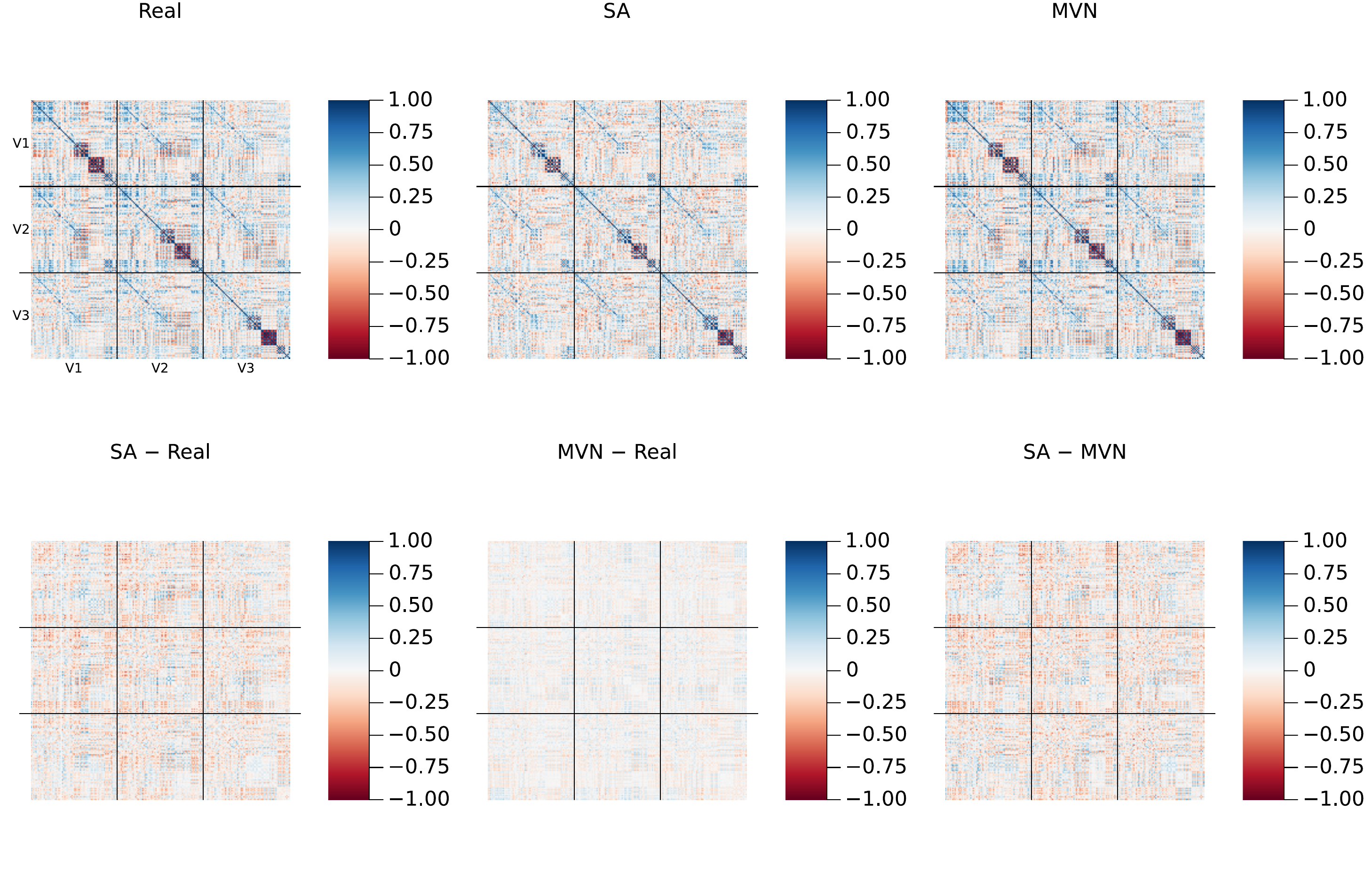}
\caption{\textbf{Cross-visit correlation structure.} Top row: full $216 \times 216$
Pearson correlation matrices for Real ($K{=}23$, left), SA ($N{=}100$, center),
and MVN ($N{=}100$, right) populations. Each matrix is organized as a
$3 \times 3$ grid of $72 \times 72$ blocks (delineated by black lines), where
the diagonal blocks capture within-visit feature correlations (V1--V1, V2--V2,
V3--V3) and the off-diagonal blocks capture cross-visit dependencies (e.g.,
V1--V3 correlations between baseline and third-trimester Factor~VIII).
Stochastic attention preserves both the within-visit and cross-visit block
structure. Bottom row: pairwise residual matrices (element-wise subtraction),
all plotted on the same $\pm 1$ color scale as the top row. The SA$-$Real
residual shows small, unstructured deviations. The MVN$-$Real residual is
smoother, particularly in the off-diagonal blocks, reflecting the Ledoit--Wolf
regularization that systematically shrinks cross-visit correlations toward zero.
The SA$-$MVN residual shows the largest differences in these
off-diagonal blocks, where SA retains longitudinal dependencies that MVN's
rank-deficient covariance estimate cannot represent. A version with amplified
residual color scale ($\pm 0.5$) and Frobenius norms is provided in
Fig.~\ref{sfig:cross-visit-corr-supp}.}
\label{fig:cross-visit-corr}
\end{figure}

Having established that SA reproduces marginal distributions and pairwise
relationships, we next asked whether it preserves the higher-order joint structure
across visits. We compared the full $216 \times 216$ cross-visit correlation matrices for real,
SA-generated, and MVN-generated populations. The real data exhibited a characteristic
block structure with strong within-visit correlations along the diagonal and
structured off-diagonal blocks reflecting cross-visit dependencies; for example,
a patient's Visit~1 Factor~X level predicting their Visit~3 Factor~X level. We found
that SA preserved this block structure, including the off-diagonal cross-visit
correlations (Fig.~\ref{fig:cross-visit-corr}). The SA$-$Real residual matrix showed
small, unstructured deviations distributed throughout, whereas the MVN$-$Real residual
was notably smoother, reflecting the regularization-induced shrinkage of off-diagonal
correlations toward zero. This difference was most apparent in the off-diagonal blocks,
where MVN systematically underestimated cross-visit dependencies that SA retained.
We note that Pearson correlation captures linear associations; Spearman rank
correlations, used in the mechanistic validation, would additionally capture
monotonic nonlinear dependencies. Within the linear analysis used here, stochastic attention
retained the observed cross-visit structure more closely than the regularized MVN baseline.

We examined the eigenvalue spectrum of the sample covariance matrix to understand the
structural origin of this difference (Fig.~\ref{sfig:eigenvalue-spectrum}). The real data had
rank~22, reflecting the $K{=}23$ patient constraint. Stochastic attention and MVN make different
trade-offs in handling this rank deficiency: SA truncates via PCA (zeroing variance
beyond component~18), while MVN regularizes via Ledoit--Wolf shrinkage (providing a
full-rank estimate by inflating eigenvalues in the 194 null dimensions). The
consequence of MVN's approach is that it introduces variance in dimensions where the
data contain no signal, which manifested as increased dispersion in PCA projections.
Visit-specific PCA projections showed this dispersion gap (Supplementary Fig.~\ref{sfig:pca-by-visit}).
Patients generated by stochastic attention occupied the same region of PCA space as real
patients across all three visits, while MVN-generated patients showed substantially greater scatter,
particularly at Visits~2 and~3. The spectrum and projections therefore linked the MVN
baseline's greater scatter to variance added in directions unsupported by the cohort.

\subsection{Robustness to Sample Size, Dimensionality, and Nonlinearity}

Results from a single cohort of $23$ patients could not show how stochastic attention
would perform if the amount or structure of the available data changed. We therefore asked three
questions in controlled tests (Methods, Sec.~\ref{sec:robustness-benchmarks}). First,
could generated profiles recover the true covariance? We simulated datasets in which the
covariance was known. We varied the size of the simulated datasets from $8$ to~$80$ profiles,
the number of variables per profile from $30$ to~$216$, and the complexity of the correlation patterns.
From each dataset, we constructed a stochastic-attention memory matrix. For comparison, we
estimated a Gaussian generator from the dataset's sample mean and covariance. Each generator
then produced $100$ profiles. We compared the covariance of each generated cohort
with the known population covariance. Stochastic attention had lower error in $35$ of the $39$ settings in which the
simulated dataset contained fewer profiles than variables (Fig.~\ref{fig:sim-recovery}A). For
simulations matching the clinical dataset in this study in terms of size, with $23$ profiles and $216$ variables per
profile, the Gaussian generator's covariance error was approximately $1.5$ times the stochastic-attention error
after averaging across the three levels of correlation-pattern complexity.
Second, we examined how well stochastic attention preserved characteristics of the full
$23$-patient cohort when fewer patients were available. Using random subsets of $20$, $15$, $10$, or~$8$ patients, we repeated the random selection ten times
at each size, constructed a new memory matrix from each, generated $100$ profiles, and compared every generated cohort with the same full $23$-patient
reference. As the subset size decreased from $20$ to~$8$, error in the cross-visit correlations
increased from $0.63$ to $1.11$ (Fig.~\ref{fig:sim-recovery}B), the median error in variable
means increased from $2.0\%$ to $4.9\%$, and overlap with the mechanistic-model
reference remained near $0.90$. Third, could stochastic attention and the Gaussian generator
reproduce curved relationships? We used a noisy S-curve, a simulated path that resembled the
letter S because it had two opposing bends. We spread this path across either $30$ or~$120$
variables while preserving its shape (Methods, Sec.~\ref{sec:robustness-benchmarks}). For each amount of curvature and each number of variables, we created five simulated
datasets, each with $23$ profiles. From every dataset, we constructed a stochastic-attention
memory matrix and estimated a Gaussian generator; each generator produced $100$ profiles. As
curvature increased, stochastic-attention profiles moved farther from the curve. At curvature
$0.5$, their mean distance was at least twice the straight-line value after averaging across
the five repeats and both numbers of variables. It remained below this threshold at curvature
$0.25$. Despite this decline, at every nonzero curvature, stochastic-attention profiles
remained closer to the curve on average than Gaussian profiles. However, the Gaussian
generator reproduced the population covariance more accurately than stochastic attention for
the curved data. Together, these tests showed that stochastic attention recovered covariance more accurately than the Gaussian
comparator in most of the small linear simulations with more variables than profiles, but its
clinical correlations and variable means became less accurate when fewer patients were used.
On curved data, each generator performed better on a different measure.

\subsection{Interpolation and Extrapolation}

The robustness tests examined recovery under controlled changes in the data. We next
asked whether profiles generated from the clinical cohort remained within the real cohort or
extended beyond it. Every synthetic profile lay outside the convex hull of the $23$ real profiles in the
$d{=}18$ PCA space (Supplementary Table~\ref{stab:hull-geometry}). However, each real profile
also lay outside the hull of the other $22$, so binary membership was uninformative in this
sparse space. We therefore compared hull distances against $22$ real profiles. Each real
calculation omitted the profile being evaluated, while each synthetic calculation omitted its
nearest real profile. If generation extrapolated unusually far, synthetic distances would
exceed leave-one-out real distances. The medians were $11.0$ and $11.9$, respectively, and a
descriptive Mann--Whitney test did not detect a difference ($p{=}0.07$). While this did not establish
equivalence, it provided no evidence that synthetic profiles extended farther from the
cohort than real profiles treated as unseen. Geometric similarity did not establish biological
plausibility, so we applied two further checks. The first required nonnegative values where
appropriate and limited coagulation-factor and TGA values to five real-cohort standard
deviations from the mean. Fifty-four of $100$ synthetic
profiles passed; the other $46$ contained a negative estradiol or progesterone value. The
unconstrained multivariate-normal baseline produced negative hormones in $43$ profiles.
Neither method enforced positivity, and positive factors in these draws did not guarantee
positivity in others. The second check used the BZ2012 model to test the generated factor combinations.
Estradiol and progesterone were not model inputs, and all nine coagulation factor inputs were positive, so
all $100$ profiles were included. For each TGA feature, $83$--$92\%$ of profiles had a
predicted-to-recorded ratio within the real 5th--95th percentile range. Forty-one met this
criterion for every feature and visit, and $24$ passed both checks. Positive-variable
transformations or constrained decoding could prevent negative values. Together, these checks
separated geometric extrapolation from biological plausibility. Hull distance provided no
evidence that synthetic profiles extended farther than real profiles treated as unseen, but
only $24$ of $100$ profiles passed both the biological and strict mechanistic criteria. Thus,
hull position alone could not determine whether an individual generated profile was
biologically plausible.

\subsection{Membership-Inference Risk}

Geometric proximity raised a separate privacy question: could a synthetic cohort reveal
whether an already-known de-identified assay profile had been stored in the memory matrix?
To answer this question, we tested whether distance from a candidate profile to the synthetic
cohort could distinguish profiles used to construct the memory matrix from profiles withheld
from it. In each of $40$ random splits of the $23$-patient cohort, we constructed the memory
matrix from $15$ profiles, held out the other $8$, and generated a synthetic cohort. A smaller
distance to the nearest synthetic profile provided stronger evidence that a candidate had
been stored. Stored profiles were closer to the synthetic cohort than held-out profiles
(median $10.7$ versus $15.4$; Supplementary Table~\ref{stab:membership}). We quantified this
discrimination using the area under the receiver operating characteristic curve (AUC). The
AUC was the probability that the test ranked a randomly chosen stored profile as more likely
to have been stored than a randomly chosen held-out profile; $0.5$ indicated chance
discrimination and $1.0$ indicated perfect discrimination. The mean AUC was $0.97$, and nine
splits reached $1.0$. We next asked whether this signal came from the numerical sampling
procedure. A Metropolis-adjusted procedure left the mean AUC nearly unchanged at $0.96$.
Lowering the inverse temperature over a $40$-fold range increased the median distance from
synthetic profiles to their nearest stored profiles, but cross-visit covariance fidelity
declined. Drawing directly from the target distribution, without a Markov chain, produced a
mean AUC of $0.98$ (Supplementary Table~\ref{stab:sampling-ladder},
Fig.~\ref{sfig:sampling-ladder}). Thus, the membership signal came from the target
distribution in this small-cohort setting, not from the Markov-chain implementation. The test
could reveal whether an already-known de-identified profile had been stored, but it could not
identify a person or reconstruct an unknown record.

\subsection{Conditional Generation of Rare Subgroups}
\label{sec:results-l3}

The previous analyses used unconditioned SA, generating from the full cohort. A key
clinical application, however, is generating patients from specific subpopulations
that are too small to study independently. We therefore tested whether SA could
amplify rare clinical subpopulations while preserving their
condition-specific signatures. We defined three overlapping subgroups by pregnancy
outcome and comorbidity: Uncomplicated ($n{=}18$), PCOS ($n{=}3$, cross-cutting;
1~patient also in the PE group), and Developed~PE ($n{=}5$). The PCOS and
Developed~PE groups were too small for any independent statistical analysis. We used
SA's multiplicity-weighted sampling to generate condition-specific cohorts of 100
patients each by upweighting attention on the relevant stored patterns.

\begin{figure}[p]
\centering
\includegraphics[width=\textwidth]{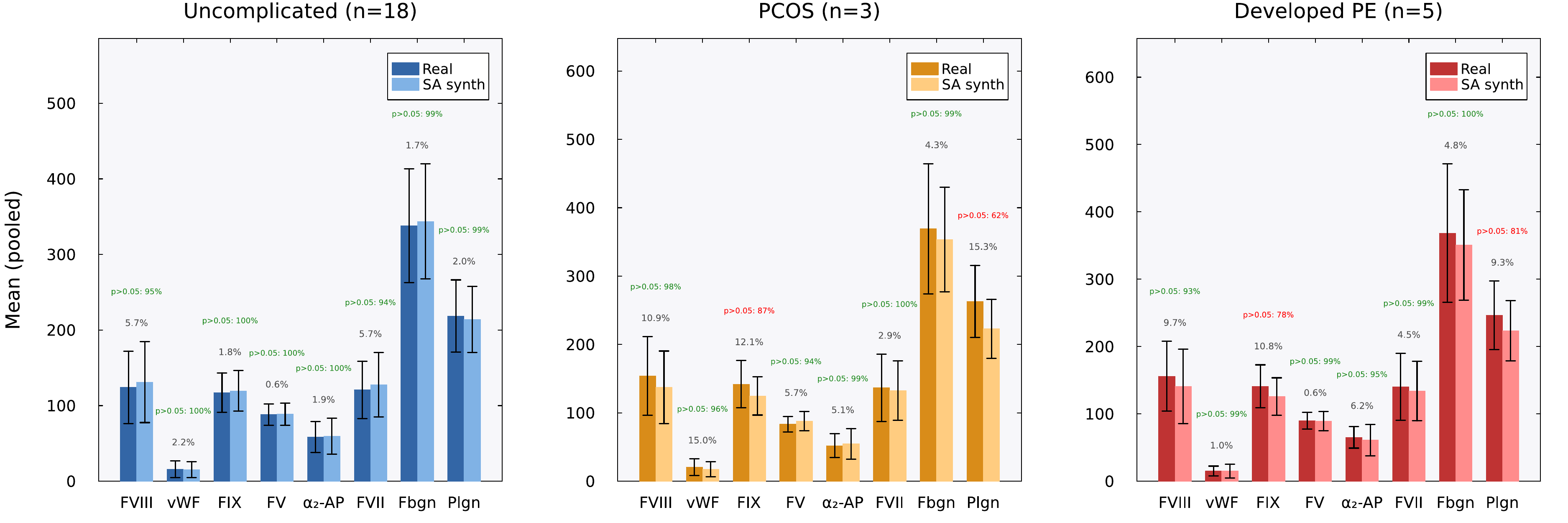}
\caption{\textbf{Condition-specific feature preservation.} Grouped bar charts
comparing real (dark bars) and SA-generated synthetic (light bars) patient
means ($\pm$ SD error bars) for eight coagulation features, shown separately
for each clinical subgroup: Uncomplicated ($n{=}18$, left, blue), PCOS ($n{=}3$,
center, orange), and Developed~PE ($n{=}5$, right, red). Each bar is the mean of
that assay pooled across all three visits (V1--V3); factor activities (FVIII,
vWF, FIX, FV, $\alpha_2$-AP, FVII, plasminogen) are expressed as \% of normal
pooled reference plasma, and fibrinogen (Fbgn) in mg/dL. Gray percentages
indicate the mean relative error (MRE). Green
annotations show a descriptive bootstrap Mann--Whitney diagnostic: the fraction
of 1,000 bootstrap replicates (synthetic subsampled to $n_{\text{real}}$) in
which $p > 0.05$, meaning that this test detected no difference at the available
sample size. This is not an equivalence test.
Features achieving $\geq$90\% are shown in green; $<$90\% in red. Across all
24 feature--condition pairs, 20 (83\%) had no difference detected in
$\geq$90\% of replicates (median: 98.6\%). The four pairs below this
threshold were FIX and plasminogen in the PCOS and Developed~PE subgroups,
where small subgroup sizes and large inter-patient variability reduced test power.}
\label{fig:conditioned-features}
\end{figure}

We compared condition-specific feature means between real and SA-generated patients
across eight coagulation features that exhibited between-condition variation
(Fig.~\ref{fig:conditioned-features}). The generated cohorts preserved the
condition-specific patterns. Patients with PCOS showed elevated Factor~VIII and vWF
relative to Healthy patients in both real and synthetic data, PE patients showed
elevated $\alpha_2$-AP and Factor~IX, and the between-condition rank ordering was
maintained across all eight features. We then performed a descriptive bootstrap
Mann--Whitney diagnostic. For each feature and condition, we subsampled the
synthetic cohort to match the real sample size ($n_{\text{real}}$), applied a
Mann--Whitney U test, and repeated 1,000 times. We reported the fraction of
replicates where $p > 0.05$, meaning that this test did not detect a difference in
that replicate. This power-limited fraction was descriptive and did not establish
equivalence. Across all 24 feature--condition pairs, 20 (83\%) achieved
$p > 0.05$ in $\geq$90\% of replicates, with a median
no-difference-detected fraction of 98.6\% (full per-pair results in
Table~\ref{stab:bootstrap-mw}). The four pairs that fell below 90\% were
high-variance features in the smallest subgroups (FIX and plasminogen in PCOS;
FIX and plasminogen in Developed~PE), where the real data exhibited large
inter-patient variability. No multiple-testing correction was applied because this
diagnostic was not used to make simultaneous inferential claims; the per-pair values and the
90\% threshold were descriptive reporting summaries. The PCA projections of the conditioned cohorts are provided in
Figs.~\ref{sfig:conditioned-pca}--\ref{sfig:conditioned-by-visit}. These results showed that SA's multiplicity-weighted
interpolation can amplify rare subgroups in a way that preserves their distinguishing
clinical features, useful for exploratory hypothesis generation and simulation-based
workflow testing, but not for increasing the effective clinical sample size. The multivariate-normal model cannot do this:
fitting a separate MVN to three PCOS patients across 216 features is mathematically
impossible (rank~2 covariance), and post-hoc subsetting of unconditional MVN samples
does not preserve subgroup-specific structure. Together, these comparisons showed that
multiplicity weighting preserved the subgroup patterns represented in the source cohort.

Generating $100$ conditioned profiles did not add independent subgroup information.
The PCOS and Developed~PE signals still came from three and five real patients, and their
participation ratios were $4.6$ and $7.7$ (Table~\ref{stab:multiplicity}). Thus, repeated draws
were concentrated around a few stored profiles. Treating them as independent observations
would understate uncertainty. They instead provided denser simulations of the distributions
suggested by those patients for exploratory modeling and workflow testing.

\subsection{Mechanistic Consistency}
\label{sec:results-l4}

\begin{figure}[p]
\centering
\includegraphics[width=\textwidth]{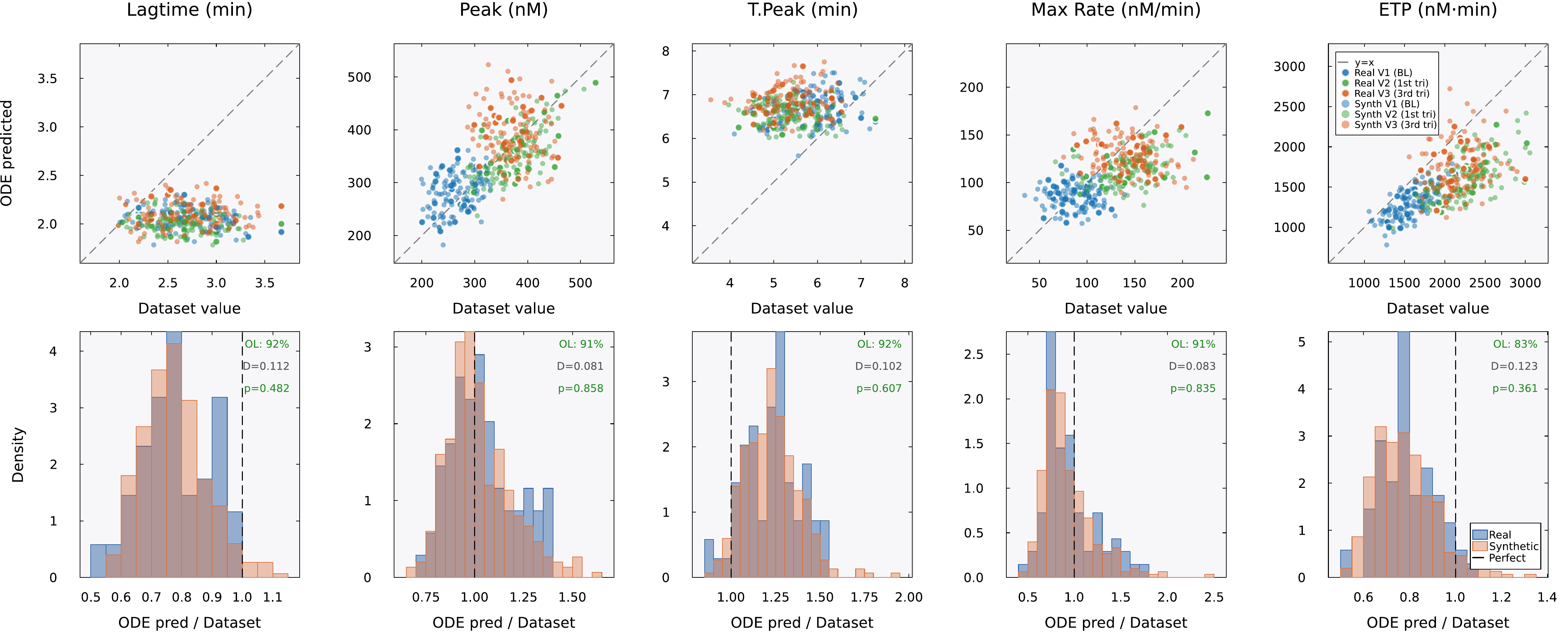}
\caption{\textbf{Mechanistic validation (TF-only).} Top row: BZ2012
ODE-predicted TGA values (vertical axis) versus dataset TGA values (horizontal
axis) for five thrombin generation parameters. The dataset values are the TGA
measurements from each patient's record, present for both real and synthetic
patients. The ODE-predicted values are computed by running each patient's
coagulation factor levels through the 58-species BZ2012 model. The dashed line
indicates where the ODE model would perfectly reproduce the dataset value
($y{=}x$); systematic deviations from this line (e.g., lagtime underprediction)
reflect ODE model bias, not synthetic data failure. Filled markers are real
patients, open markers are SA-generated synthetic patients, colored by visit:
blue (V1/baseline, training set), green (V2/first trimester), orange (V3/third
trimester). Real and synthetic patients occupy the same cloud with the same
pattern of ODE model bias. Bottom row: ODE-predicted/dataset ratio
distributions for real (blue) and synthetic (orange) patients. The
distributions overlap substantially. Each panel is annotated with cloud overlap
(OL), the two-sample Kolmogorov--Smirnov statistic ($D$), and $p$-value. All
five features had $p>0.35$; these tests did not detect a difference between real and
synthetic ratio distributions.}
\label{fig:mechanistic-combined}
\end{figure}

The preceding validations assessed statistical properties of the synthetic data. We
next tested whether factor combinations generated by stochastic attention produced
biologically plausible outputs when passed through a separately specified mechanistic model.
The model was blind to the generation process but was calibrated on the same cohort.
We used the Hockin--Mann BZ2012 model~\citep{Hockin2002,BrummelZiedins2012}, a system
of 58 ordinary differential equations with 64 rate constants that simulates thrombin
generation from patient-specific coagulation factor inputs. We calibrated 5 of the
64 rate constants (prothrombinase, intrinsic and extrinsic tenase, protein~C
activation, and meizothrombin conversion; Table~\ref{stab:ode-calibration}) on Visit~1 real
patients at the population level and held the remaining 59 at published literature
values. We then ran the calibrated model on all 23 real patients and 100 synthetic
patients under both TF-only and TF+TM conditions (738 total simulations, 0 failures). For each patient, we
computed five ODE-predicted thrombin generation features (lagtime, peak, time-to-peak,
maximum rate, and ETP) and compared them against the corresponding values from the
patient record. In these comparisons (Fig.~\ref{fig:mechanistic-combined}, top row),
the horizontal axis is the TGA value from the patient record, which exists for both
real and synthetic patients, while the vertical axis is the value computed by the
BZ2012 model from that patient's factor inputs. Systematic deviations from the
$y{=}x$ line reflect ODE model bias (e.g., lagtime is underpredicted at
$\approx$0.77$\times$), not a failure of the synthetic data; the key observation is
that real and synthetic patients share the same pattern of model bias, occupying the
same cloud with the same systematic offset.

We formalized this observation by computing the ODE-predicted/dataset ratio for each
patient and comparing the resulting distributions between real and synthetic
populations (Fig.~\ref{fig:mechanistic-combined}, bottom row). These ratio
distributions capture how the ODE model processes each patient's factor levels. If
SA had generated patients with biologically implausible factor combinations, the
model would process them differently, producing a shifted or broadened ratio
distribution. Instead, the distributions overlapped substantially, with cloud
overlap (fraction of synthetic ratios within the 5th--95th percentile of real
ratios) ranging from 0.83 for ETP to 0.92 for T.Peak under TF-only conditions,
and two-sample Kolmogorov--Smirnov (KS) tests did not detect a difference between
the real and synthetic ratio distributions at this sample size across all five TGA features
($D = 0.081$--$0.123$, all $p > 0.30$; full diagnostics in
Table~\ref{stab:mechanistic-validation}).
Under TF+TM conditions, the model systematically overcorrected the protein~C
anticoagulant pathway, producing peak and maximum rate predictions
approximately 0.5$\times$ measured values for both real and synthetic patients
(Fig.~\ref{sfig:mechanistic-tm-combined}, same layout as the main-text
comparison). Cloud overlap remained high (0.88--0.92 across the
five features), showing similar model behavior for real and synthetic profiles despite the
imperfect calibration.

The model was calibrated only on Visit~1 real patients, but the per-visit median predicted-to-measured
ratios under TF-only conditions remained close to 1.0 at Visits~2 and~3
(Fig.~\ref{sfig:generalization}). This was consistent with rate constants that remained useful
beyond the calibration visit. Spearman rank correlations between predicted and measured values were
moderate for ETP ($\rho \approx 0.6$--$0.8$ for real, $\rho \approx 0.6$--$0.65$
for synthetic; Fig.~\ref{sfig:rankcorr}) and weak for other features, consistent with
a 5-parameter population-level calibration that was not designed to resolve
inter-patient variability. The key finding was not patient-level prediction but
population-level plausibility. Applied identically to both groups, the mechanistic model
produced overlapping predicted-to-recorded ratio distributions for real and synthetic patients
under both experimental conditions. Because the model was calibrated on this same cohort, this
was a same-cohort consistency check rather than independent validation.

\subsection{Downstream Utility: Mechanistic Model Calibration}
\label{sec:results-utility}

\begin{figure}[p]
\centering
\includegraphics[width=\textwidth]{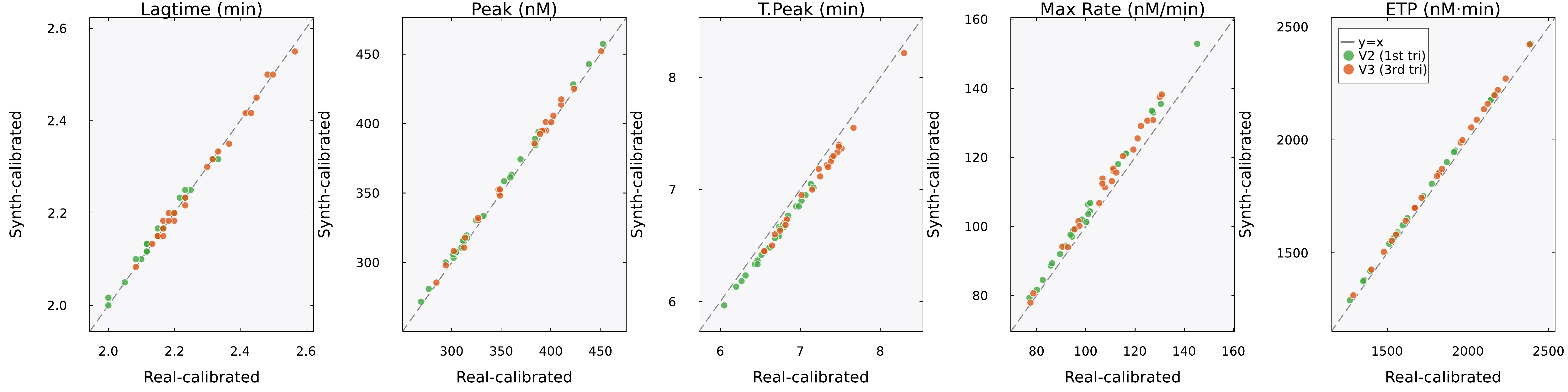}
\caption{\textbf{Downstream utility: synthetic and real calibrations on held-out real
profiles.} Each panel compares
the BZ2012 ODE predictions for one TGA feature when the model is calibrated on
real V1 patients (horizontal axis) versus synthetic V1 patients (vertical
axis), evaluated on the same held-out real V2 and V3 patients. Points near the
$y{=}x$ line indicate similar predictions from the two calibrations.
Green: V2 (first trimester); orange: V3 (third trimester). The synth-calibrated
model had slightly lower median relative error on all five features (overall
synthetic-to-real error ratio $0.94$). Both calibrations used 11 random restarts; all starts
converged.}
\label{fig:downstream-utility}
\end{figure}

Having established same-cohort mechanistic consistency, we tested whether synthetic profiles
could support mechanistic-model calibration. We fit the
five BZ2012 rate constants under TF-only conditions to either $100$ synthetic V1 profiles or
$23$ real V1 profiles, using the same bounded Nelder--Mead optimization and 11 random restarts
(Table~\ref{stab:ode-calibration}). We evaluated both calibrations on held-out real V2 and V3
TGA measurements. Median relative errors for the synthetic calibration were 2--10\% lower
across the five features, with an overall synthetic-to-real error ratio of $0.94$
(Fig.~\ref{fig:downstream-utility}; Table~\ref{stab:downstream-utility}). The parameter
estimates differed, but the two calibrations produced highly correlated predictions. Because
the synthetic profiles came from the real cohort, this was not an independent validation. It
tested whether the representation retained enough structure for calibration.
The modest reduction in prediction error did not show that synthetic data were
superior. It may instead have reflected a more stable calibration objective produced by
sampling the represented distribution more densely. Together with the subgroup and
factor-level analyses, this result supported exploratory modeling and hypothesis generation.
Confirming those hypotheses and establishing clinical utility would require additional real
patients or independent experimental data.

\section{Discussion}
\label{sec:discussion}

This study asked whether a geometry-based method could generate synthetic profiles from a
very small longitudinal cohort while preserving statistical and modeled biological
relationships. Median feature MRE was 1.2\%, which the MVN baseline also achieved.
Stochastic attention also retained cross-visit covariance, whereas MVN introduced variance in
194 null dimensions. Conditioned generation preserved the subgroup patterns represented by
three PCOS and five Developed~PE patients. Finally, a separately specified,
generator-blind coagulation model calibrated on the same cohort placed real and synthetic
profiles in the same factor-to-thrombin-generation envelope. Cloud overlap was 83--92\%, and
the Kolmogorov--Smirnov tests did not detect differences under these conditions ($p>0.35$ for
all five TGA features).

Dense simulations from very small longitudinal datasets could support maternal health
research. Rare pregnancy complications such as PCOS and preeclampsia are difficult to study
because assembling large, well-characterized longitudinal cohorts requires years of
recruitment across multiple clinical sites. Our results support hypothesis generation,
computational modeling, and workflow testing with dense simulations of these small cohorts.
Related work has predicted TGA responses from coagulation factors and searched for
factor changes that produce a target response~\citep{Ghetmiri2021}. Although we did not
evaluate either task with profiles generated by stochastic attention, these applications
offer a clear direction for extending the present work.

Understanding why stochastic attention worked in this setting required examining the role of the
PCA representation as well as the generator. When all methods used
the same $18$-dimensional subspace, the fitted two-component mixture and copula achieved
marginal errors and mechanistic overlaps similar to those of stochastic attention. These results were therefore
not unique to stochastic attention, although the comparison did not quantify the contribution
of PCA itself. Empirical correlation error and novelty differed across generators. Stochastic
attention used the patient profiles in the PCA subspace as its memory matrix and avoided
estimating a full $216$-dimensional covariance from $23$ patients. Its sampler still required
a choice of inverse temperature. A natural concern is whether the $\beta^*$
selection procedure, originally developed for protein sequence
generation~\citep{Varner2026SAProtein}, transfers to clinical coagulation data. The entropy
transition rule selected the point of greatest downward curvature in attention entropy, a
geometric property of $K$ patterns in $d$ dimensions rather than a property of what the
features represented. For this dataset it gave $\beta^* \approx 2.94$, with the random-pattern
scale $\sqrt{d} \approx 4.24$ serving as a theoretical reference. A sensitivity analysis
showed gradual changes in generation quality across $\beta/\beta^* \in [0.1, 3.0]$
(Table~\ref{stab:beta-sweep}).
Similarly, varying the PCA variance retention threshold from 85\% to 99\%
($d_{\text{PCA}} = 13$ to 21) had minimal impact on generation quality. Median MRE
remained between 1.2\% and 1.8\% across all thresholds (Table~\ref{stab:pca-sensitivity}),
so the 95\% threshold was not a critical design choice. Together, these sensitivity
analyses showed that performance was not tied to a narrow choice of inverse temperature or
PCA threshold within the ranges tested.

Conditional generation raised a different question: how much diversity remained after the
selected subgroup was upweighted? Achieving
80\% of finite-mixture component probability on the PCOS subgroup required
$\rho \approx 26.7$, reducing the participation ratio to
$K_{\text{eff}} \approx 4.6$ (multiplicity parameters
for all three subgroups in Table~\ref{stab:multiplicity}),
which indicated that mixture-component probability was concentrated around a small number
of stored profiles across repeated draws. The subgroup-specific signal still rested on only
three real PCOS profiles, and the lower-weight background profiles pulled generated means
toward the population average; $K_{\text{eff}}$ did not represent an independent patient
count. Magnitudes were drawn from all $23$ empirical PCA-space norms, so conditioning acted
through the weighted directions rather than a subgroup-specific magnitude distribution.
Thus, weighting concentrated sampling toward the subgroup directions while retaining some
influence from the full cohort.
Related studies have applied the same geometry-based operating-point rule
and multiplicity weighting to protein sequence generation and conditional
steering~\citep{Varner2026SAProtein,Varner2026Multiplicity,Alswaidan2026SA}.

Among longitudinal clinical generators, VAMBN-MT provides a useful comparison in a
larger-data setting~\citep{Kuhnel2024}. It combines
expert-defined modules, Bayesian-network constraints, and LSTM-augmented variational
autoencoders. The authors evaluated it on the DONALD nutritional cohort and the Alzheimer's
Disease Neuroimaging Initiative. DONALD included $1{,}312$ participants and 33 longitudinal
variables, compared with 23 patients and 216 features here. Its best variant had cross-visit
correlation error near 0.70, and weaker time trends were not reproduced at the tested sample
sizes. The methods address different regimes. The VAMBN-MT model uses a trained modular architecture
for moderate cohorts. Stochastic attention uses PCA geometry when $n<p$, applies a generic
generator-blind mechanistic check instead of domain-specific rules, and uses multiplicity
weights for conditional generation. This was a regime comparison, not a head-to-head test.

The mechanistic analysis addressed a different validation question: whether a model of the
underlying biology processed real and synthetic profiles similarly. We processed both groups
with the same separately specified model, which was blind to the generator but
calibrated on the real cohort. A model calibrated on synthetic profiles also
predicted held-out real TGA measurements as well as one calibrated on real profiles (error
ratio $0.94$, with 11 random restarts). The same comparison can be used wherever a validated
mechanistic model exists, including pharmacokinetic~\citep{MouldUpton2013},
physiological~\citep{Chase2011}, tumor-growth~\citep{Ribba2012}, and
metabolic~\citep{Yizhak2015,Shan2018} models. The relevant test is whether the model places
real and synthetic profiles in the same predicted envelope.

The subgroup analysis focused on average differences between groups, but it did not
show whether the conditioned cohorts preserved clinically important extremes. In the
descriptive bootstrap Mann--Whitney diagnostic, the test detected no difference in fewer
than 90\% of replicates for four of the 24 feature--condition pairs: FIX and plasminogen in
the PCOS and Developed~PE groups. For these pairs, the relative differences between the real
and synthetic subgroup means were 9--15\%, and values varied widely among the real patients.
This diagnostic had limited power and was not an equivalence test. The lower fractions could
therefore reflect shifts in the means, a loss of extreme values, or simply the small samples.
Truncating the PCA representation and drawing profile magnitudes from only $23$ observed
values could also make the generated extremes too narrow. If so, a synthetic cohort could
not show how often severe states occur or how severe they become. Profiles outside the real cohort's
convex hull did not show that the true extremes were reproduced, and the primary mechanistic
cloud-overlap measure considered only the real 5th--95th percentile range. We therefore do
not recommend using these cohorts to estimate extreme risk without additional real data.
The small source cohort caused other limits. Hull membership had little resolution,
and the conditioned cohorts still relied on only three PCOS or five Developed~PE patients.
The modest calibration improvement was consistent with drawing more profiles from the same
patient-derived distribution, not with adding biological information. The generated profiles
supported simulation and model testing, but they did not add clinical evidence beyond the
source cohort.

The pipeline required every patient to have the same assays at the same aligned visits.
Concatenating the visits preserved relationships across visits, but the pipeline could not
directly use irregular visit times, missing assays or visits, or different numbers of visits.
Imputation, padding, or a sequence-aware method could support these records, but each would
add assumptions that require validation in a small cohort.
PCA added another limit because it captured only linear patterns. In the S-curve
benchmark, generated profiles moved farther from the known curve as curvature increased.
At every nonzero curvature, stochastic attention remained closer to the curve on average
than the Gaussian generator, whereas the Gaussian generator recovered covariance more
accurately. Neither generator was best on both measures. A nonlinear representation could
better follow curved or branching patterns, but it would add choices about fitting the model
and converting the generated values back to patient measurements. Those choices are
difficult to make from $23$ patients. The linear representation was practical for this
cohort, but the results did not show that clinical trajectories are generally linear.
Cohort size also affects computational cost. The reported Langevin sampler required
work proportional to $TKd$ for each profile, where $T$ was the number of Langevin steps,
$K$ was the number of stored profiles, and $d$ was the dimension of each profile in the
sampling space. The exact mixture sampler removed the $T$ sequential updates: its component
weights could be prepared once, after which each profile
required a component draw, a $d$-dimensional Gaussian draw, and conversion back to the
original variables. The simulation advantage over the sample-covariance Gaussian occurred
mainly in the $n<p$ settings, and we did not test cohorts larger than $23$ patients.
The results therefore did not establish performance on irregular, incomplete, or
strongly nonlinear records or in larger clinical cohorts.

This study had two broad limitations: the absence of independent clinical data and the
restricted scope of the validation analyses. First, the clinical analysis used one
longitudinal cohort of $23$ patients. The known-covariance simulations tested selected
properties under controlled conditions, and related protein studies applied the sampling
framework in another domain, but neither established that the clinical findings would
generalize to a different patient population. We did not test the method in a second clinical
cohort, which is needed to assess that generalizability. Second, the validation analyses
answered narrower questions than full biological or clinical validity. The BZ2012
model was calibrated on the same cohort used for generation. Applying one fixed,
generator-blind model to real and synthetic profiles therefore tested consistency with that
model, not independent biological validity.
The fitted rate constants also required a 50-fold reduction in intrinsic tenase
$k_{\text{cat}}$ and a 15-fold
increase in extrinsic tenase $k_{\text{cat}}$ relative to published values
(Table~\ref{stab:ode-calibration}), likely because the published measurements and the
plasma-based assay used different experimental conditions. These fitted values should not be
read as new biochemical estimates. The biological scope was also incomplete. The
mechanistic test covered thrombin generation but not the fibrinolytic or viscoelastic
features. The membership-inference analysis answered a separate, narrow question: whether an
already-known de-identified profile had been stored in the memory matrix. It distinguished
stored from held-out profiles, but it did not identify a participant or reconstruct an
unknown record. Finally, we did not test clinical outcomes. Thus, the results
did not establish biological fidelity across the full profile or clinical utility. An
independently calibrated model, models covering the other feature classes, and prospective
outcome validation could address these gaps. The results establish proof of concept, not
generalizability.

\section{Conclusion}
\label{sec:conclusion}

Multiplicity-weighted Stochastic Attention generated longitudinal profiles from $23$ patients
with low marginal error, retained cross-visit covariance, and preserved the represented
subgroup patterns. The profiles were also consistent with a coagulation ODE model
calibrated on the same cohort. A model calibrated on synthetic profiles predicted held-out
real TGA measurements as well as one calibrated on real profiles. Direct sampling from
the same target distribution reproduced the Langevin sampler's fidelity and
membership-inference results. Together, these findings provide proof of concept for
using synthetic profiles in the modeling tasks tested here with very small longitudinal
cohorts.

\section{Methods}
\label{sec:methods}

\subsection{Population Dataset}
\label{sec:dataset}
We analyzed a longitudinal dataset of coagulation, fibrinolysis, thrombin generation
assay (TGA), and rotational thromboelastometry (ROTEM) measurements from women
desiring a pregnancy. The study was approved by the Institutional Review Board at
the University of Vermont. All participants gave written consent before enrollment.
Blood collection, plasma processing, and assay protocols have been described
previously~\citep{McLean2012TGA,Hale2012,Psoinos2021,Bernstein2016}. Briefly, citrated platelet-poor plasma
was collected without tourniquet use after supine rest at three clinical visits:
Visit~1 (pre-pregnancy baseline in the follicular phase), Visit~2 (end of first
trimester), and Visit~3 (mid-third trimester). Plasma aliquots were stored at
$-80^{\circ}$C for subsequent analysis. Measurements of secondary analytes were
measured as follows: Stago clotting assays measured coagulation-factor activity, ELISA
measured fibrinolysis proteins, and the CLIA-certified laboratory at the University of Vermont
Medical Center measured hormone levels. We performed TGA using 5~pM relipidated tissue factor with
fluorogenic substrate on a Synergy4 plate reader as described
in McLean et al.\citep{McLean2012TGA}. Clot formation and fibrinolysis dynamics were assessed via
viscoelastometry using the ROTEM Delta Instrument (Werfen). Thawed plasma was mixed
with or without 4~nM tPA (with respect to plasma volume), recalcified with 15~mM
calcium chloride, and incubated for 3~minutes at 37$^{\circ}$C. The reaction was
initiated by adding the plasma mixture to ROTEM cups containing tissue factor at a
17:1 ratio; the final concentration of TF in the reaction was 5~pM.

The dataset contained 153 total records across approximately 50 subjects. After
removing columns with $>30\%$ missing values and retaining only subjects with
complete data across all three visits, we obtained $K=23$ complete longitudinal
profiles across $n=72$ assay measurements per visit.
The 72 assay measurements spanned five categories: (i)~hormones (estradiol,
progesterone), (ii)~coagulation factors and inhibitors (Factors~II, V, VII, VIII,
IX, X, XI, XII; antithrombin (AT), protein~C (PC), tissue factor pathway inhibitor (TFPI),
von Willebrand factor (vWF), fibrinogen, etc.), (iii)~fibrinolytic markers
(plasminogen, plasminogen activator inhibitor-1 (PAI-1), plasminogen activator inhibitor-2 (PAI-2),
$\alpha_2$-antiplasmin ($\alpha_2$-AP), thrombin-activatable fibrinolysis inhibitor (TAFI),
tissue plasminogen activator (tPA), D-dimer equivalents),
(iv)~thrombin generation parameters under four initiator conditions (TF at 5~pM,
TF+thrombomodulin (TM), No~TF, No~TF+anti-FXIa), and (v)~ROTEM/viscoelastic parameters
(clotting time (CT), maximum clot firmness (MCF),
alpha angle, lysis onset time (LOT), maximum lysis (ML), lysis time (LT),
area under the curve (AUC) under TF~Only and TF+tPA conditions).

Demographics and clinical characteristics of the resulting cohort, including
age at enrollment, prepregnancy BMI, parity, gestational ages at each visit,
and the cross-tabulation of enrollment cohort against pregnancy outcome, are
summarized in Table~\ref{tab:demographics}. The 23 patients were drawn from
three enrollment conditions: Healthy nulliparous women ($n{=}14$), women with
a personal history of preeclampsia from a previous pregnancy (Prior~PE,
$n{=}6$), and women with polycystic ovarian syndrome (PCOS, $n{=}3$;
2~individuals were nulliparous). For conditional generation, we defined three
overlapping subgroups based on pregnancy outcome and comorbidity rather than
enrollment condition: Uncomplicated ($n{=}18$, all patients whose study pregnancy
did not result in preeclampsia), PCOS ($n{=}3$, a cross-cutting comorbidity;
1~patient also developed PE), and Developed~PE ($n{=}5$, patients who developed
preeclampsia during the study pregnancy, drawn from 2~healthy-enrolled,
2~prior-PE-enrolled, and 1~PCOS-enrolled participants).

\subsection{Multiplicity-Weighted Stochastic Attention}
\label{sec:sa}

We generated synthetic patients using a multiplicity-weighted variant of Stochastic
Attention (SA) rooted in modern Hopfield network theory~\citep{Ramsauer2021}. We
stored $K$ memory patterns as columns of a matrix $\mhat{X}\in\R^{d\times K}$ and
defined a weighted Hopfield energy landscape:
\begin{equation}
    E_r(\vxi) = \frac{1}{2}\norm{\vxi}^2
    - \frac{1}{\beta}\log\sum_{k=1}^{K} r_k \exp\left(\beta\,\mathbf{m}_k^\top\vxi\right)
    \label{eq:hopfield_energy}
\end{equation}
where $\{\mathbf{m}_k\}_{k=1}^K$ are stored memory patterns, $\beta$ is an inverse
temperature parameter controlling the sharpness of retrieval, and $\{r_k\}_{k=1}^K$
are per-pattern multiplicity weights. When all $r_k = 1$, this reduces to the standard
modern Hopfield energy; when $r_k = \rho > 1$ for a designated subset of patterns, the
energy landscape is continuously deformed to favor that subset.

This energy defines a probability distribution that can be written exactly. Writing
$\pi_r(\vxi)\propto\exp[-\beta E_r(\vxi)]$ and completing the square in each exponent gives
the expression:
\begin{equation}
    \exp\!\left[-\frac{\beta}{2}\norm{\vxi}^2 + \beta\,\mathbf{m}_k^\top\vxi\right]
    = \exp\!\left(\frac{\beta}{2}\norm{\mathbf{m}_k}^2\right)
      \exp\!\left[-\frac{\beta}{2}\norm{\vxi-\mathbf{m}_k}^2\right],
    \label{eq:complete_square}
\end{equation}
Substituting Eq.~\eqref{eq:complete_square} into the probability density gives an exact
finite isotropic Gaussian mixture:
\begin{equation}
    \pi_r(\vxi) = \sum_{k=1}^{K} q_k\,
    \mathcal{N}\!\left(\mathbf{m}_k,\ \beta^{-1}\mathbf{I}\right),
    \qquad
    q_k \propto r_k\exp\!\left(\frac{\beta}{2}\norm{\mathbf{m}_k}^2\right).
    \label{eq:exact_mixture}
\end{equation}
Each stored patient is the center of one component. The common component variance is
$\beta^{-1}$, and the multiplicity weights enter the mixture probabilities directly. The
memories used here are normalized to unit length, so the exponential factor is shared by
every component and $q_k = r_k/\sum_j r_j$ exactly. We could therefore sample this
distribution directly: first choose $k$ from $q$ and then draw
$\vxi\sim\mathcal{N}(\mathbf{m}_k,\beta^{-1}\mathbf{I})$. This procedure requires no Markov
chain. The cohort
analyzed in this work was generated with the Langevin implementation given next, and that is
what the reported results describe. We added the direct sampler for the stochastic-attention
target afterward as a check,
using the same normalization, magnitude rescaling, and decoding steps
(Table~\ref{stab:sampling-ladder}).

We sampled from this landscape using an Unadjusted Langevin Algorithm (ULA):
\begin{equation}
    \vxi_{t+1} = (1-\alpha)\,\vxi_t
    + \alpha\,\mhat{X}\,\text{softmax}\!\left(\beta\,\mhat{X}^\top\vxi_t + \log\mathbf{r}\right)
    + \sqrt{\frac{2\alpha}{\beta}}\,\boldsymbol{\varepsilon}_t
    \label{eq:ula}
\end{equation}
where $\alpha$ is a step size, $\boldsymbol{\varepsilon}_t\sim\mathcal{N}(\mathbf{0},\mathbf{I})$,
and $\log\mathbf{r}$ is the element-wise log of the multiplicity vector. The
deterministic component drove $\vxi_t$ toward a multiplicity-weighted combination of
stored patterns, while the stochastic component added variation. The
$\log r_k$ bias in the softmax logits interpolated between unconditioned generation ($\rho = 1$,
all patterns equally weighted) and hard subset curation ($\rho \to \infty$, only
designated patterns contribute). This is an inference-time operation that requires
no change to the stored patterns.

The synthetic cohort analyzed in this work was produced by initializing each Langevin
chain near a randomly selected stored pattern and retaining the final state after $T{=}2000$
steps. Because the starting point and chain length could affect the output, we compared four
sampling procedures. These used the original initialization near a stored pattern, random
initialization on the unit sphere, pooled samples from multiple chains after burn-in and
thinning, and Metropolis-adjusted sampling. As a separate check, we also drew directly from
the exact stochastic-attention mixture in Eq.~\eqref{eq:exact_mixture}, without using a Markov
chain. The fidelity and
membership-inference results were similar across all four procedures and the direct draws.
This agreement indicated that the original implementation reproduced the evaluated behavior
of the sampling distribution at $\beta^*$ (Supplementary
Fig.~\ref{sfig:sampling-ladder}).

The inverse temperature $\beta$ affects Hopfield retrieval and direct sampling in
different ways. In the Hopfield update, small $\beta$ spreads attention across many stored
profiles and tends toward their weighted average, whereas large $\beta$ concentrates attention on
individual profiles. When sampling directly from Eq.~\eqref{eq:exact_mixture}, the stored
profile is chosen according to $q_k$. For the unit-normalized profiles used here, these
probabilities are independent of $\beta$ and are set by the multiplicity weights. Instead,
$\beta$ controls the spread around the selected profile through the covariance
$\beta^{-1}\mathbf{I}$. Small $\beta$ therefore produces broader draws, whereas large $\beta$
produces tighter draws. The attention-entropy transition used to define $\beta^*$ marks the
shift from attention shared across profiles to attention concentrated on individual profiles.
We selected $\beta^*$ from the transition in normalized attention entropy across a
logarithmic sweep of $\beta$. Specifically, $\beta^*$ was the grid point where the entropy
curve bent downward most strongly, calculated as the most negative finite-difference second
derivative with respect to $\log\beta$ (Supplementary Methods). This choice depended on the
relative positions of the $K$ profiles in $d$ dimensions, not on what their features
represented. For this dataset
($K{=}23$, $d_{\text{PCA}}{=}18$), it gave $\beta^*\approx2.94$; the random-pattern prediction
$\sqrt{d}\approx4.24$ served as a theoretical reference. A sensitivity analysis across
$\beta/\beta^*\in[0.1,3.0]$ showed that the selected value balanced novelty and
fidelity (Table~\ref{stab:beta-sweep}). For conditioned generation, we repeated the entropy
calculation with the multiplicity bias included, giving a $\rho$-dependent $\beta^*(\rho)$.

We used the participation ratio to measure how strongly multiplicity weighting
concentrated repeated draws around a few stored profiles. For unit-normalized memories,
$q_k=r_k/\sum_j r_j$ and
$K_{\text{eff}}=1/\sum_k q_k^2=(\sum_k r_k)^2/\sum_k r_k^2$. Thus,
$K_{\text{eff}}$ is the number of equally weighted components with the same concentration.
Equal weights gave $K_{\text{eff}}=23$; concentrating weight on three profiles moved it toward
three. This was not an independent clinical sample size. To assign a target fraction
$f_{\text{target}}$ of component probability to a subgroup, we used
$\rho=f_{\text{target}}K_{\text{bg}}/[K_{\text{des}}(1-f_{\text{target}})]$, where
$K_{\text{des}}$ and $K_{\text{bg}}$ are the subgroup and background profile counts.

\subsection{Pipeline for Continuous Longitudinal Data}
\label{sec:pipeline}

Applying SA to continuous longitudinal clinical data required several adaptations
beyond the standard Hopfield sampling framework. We first concatenated each patient's
$n{=}72$ assay measurements across all three visits into a single vector of dimension
$d_{\text{concat}} = 3n = 216$, so that each stored memory pattern encoded a
complete longitudinal profile and generated synthetic patients would have internally
consistent multi-visit trajectories. We then standardized each feature to zero mean
and unit variance and applied PCA retaining 95\% of the variance, reducing
dimensionality from 216 to $d_{\text{PCA}}{=}18$. This compression served two
purposes: it removed collinearity among the 216 features and yielded a
memory-to-dimension ratio of $K/d_{\text{PCA}} \approx 1.28$, placing SA in a
favorable operating regime where the number of stored patterns exceeded the
dimensionality of the representation.

Standard SA operates on unit-norm patterns, which is appropriate for discrete data
(e.g., one-hot protein sequences) but destroys the anisotropic variance structure
of continuous clinical measurements, collapsing dispersion across all principal
components to a unit sphere. We addressed this with a direction-magnitude
decomposition. Before normalization, we recorded each pattern's PCA-space norm
$s_k = \|\mathbf{m}_k\|$; we then ran SA on unit-norm patterns to obtain a
directional sample $\hat{\vxi}$, drew a magnitude $s$ from the empirical
distribution of $\{s_k\}$, and reconstructed the rescaled sample as
$\vxi_{\text{rescaled}} = s \cdot \hat{\vxi}$. This preserved the directional
structure learned by the Hopfield energy while restoring the natural scale of
variation. The rescaled PCA vector was mapped back to the original feature space via
inverse PCA and destandardization, then split into three 72-dimensional visit records.

To generate condition-specific cohorts (e.g., PCOS-only), we set the multiplicity
$r_k = \rho$ for patterns belonging to the target subgroup and $r_k = 1$ for all
others, and sampled using the weighted Langevin update (Eq.~\ref{eq:ula}). The
multiplicity $\rho$ was computed to place a target fraction
$f_{\text{target}} = 0.80$ of the mixture-component probability on the designated subset,
while retaining 20\% probability on background patterns, via
$\rho = f_{\text{target}} \cdot K_{\text{bg}} / (K_{\text{des}} \cdot (1 -
f_{\text{target}}))$. Multiplicity parameters for all three subgroups are
listed in Table~\ref{stab:multiplicity}. For the PCOS subgroup
($K_{\text{des}}{=}3$), this required $\rho \approx 26.7$, reducing the
participation ratio to $K_{\text{eff}} \approx 4.6$. For both unconditioned and
conditioned generation, magnitudes were drawn from the empirical distribution of all $23$
PCA-space norms. Subgroup weighting affected directional sampling, not the distribution of
magnitudes.

The complete generation pipeline is summarized in Algorithm~\ref{alg:sa-pipeline}.
The full set of SA hyperparameters used for all experiments is listed in
Table~\ref{stab:hyperparams}, with key operating values $\alpha = 0.01$,
$T = 2{,}000$ Langevin iterations per sample, and random seed 42 for
reproducibility. A control experiment comparing ULA with the
Metropolis-Adjusted Langevin Algorithm (MALA) showed that the
discretization bias is negligible at this step size. Across 10 chains of
5{,}000 iterations the MALA acceptance rate was $99.9 \pm 0.03\%$, and mean
post-burn-in energies and effective sample sizes were indistinguishable
between the two samplers (Supplementary Table~\ref{tab:ula-vs-mala}). The pipeline was implemented in Julia (v1.12) using the packages listed in the code
repository. Generating 100 synthetic patients required approximately 0.1~seconds on
a standard laptop (Apple M-series), comparable to MVN sampling ($\approx$0.15~s),
because SA operates in the 18-dimensional PCA space rather than the full
216-dimensional feature space. Each synthetic patient was generated by an independent
Langevin chain (no shared state between patients); generation quality was stable
across $T \in [500, 4000]$ iterations (median MRE 1.0--1.6\%), indicating that
$T{=}2{,}000$ was sufficient for convergence.

\begin{algorithm}[ht]
\caption{Multiplicity-Weighted SA for Longitudinal Patient Generation}
\label{alg:sa-pipeline}
\begin{algorithmic}[1]
\Require Patient data $\{\mathbf{x}_k^{(v)}\}$ for $k=1,\ldots,K$ patients, $v=1,2,3$ visits
\Require Multiplicity vector $\mathbf{r}$ (all ones for unconditioned; $r_k = \rho$ for designated subset)
\Ensure $N$ synthetic longitudinal patient profiles
\State Concatenate: $\mathbf{x}_k \gets [\mathbf{x}_k^{(1)}; \mathbf{x}_k^{(2)}; \mathbf{x}_k^{(3)}] \in \mathbb{R}^{216}$
\State Standardize: $\mathbf{z}_k \gets (\mathbf{x}_k - \boldsymbol{\mu}) \oslash \boldsymbol{\sigma}$
\State PCA: $\mathbf{m}_k \gets \mathbf{W}^\top \mathbf{z}_k \in \mathbb{R}^{18}$ \Comment{95\% variance retained}
\State Record norms: $s_k \gets \|\mathbf{m}_k\|$; normalize $\hat{\mathbf{m}}_k \gets \mathbf{m}_k / s_k$
\State Select $\beta^*$ where the attention-entropy curve bends downward most strongly, including $\log\mathbf{r}$ when conditioned
\For{$i = 1, \ldots, N$}
    \State Select anchor $k_0$ from all patterns, or from the designated subset when conditioned
    \State $\hat{\vxi}_0 \gets \hat{\mathbf{m}}_{k_0}+0.01\boldsymbol{\varepsilon}_0$; normalize $\hat{\vxi}_0$
    \For{$t = 0, \ldots, T-1$} \Comment{Langevin dynamics}
        \State $\hat{\vxi}_{t+1} \gets (1-\alpha)\hat{\vxi}_t + \alpha\hat{\mathbf{M}}\,\mathrm{softmax}(\beta^*\hat{\mathbf{M}}^\top\hat{\vxi}_t + \log\mathbf{r}) + \sqrt{2\alpha/\beta^*}\,\boldsymbol{\varepsilon}_t$
    \EndFor
    \State Draw magnitude: $s \sim \{s_k\}_{k=1}^K$
    \State Rescale: $\mathbf{m}_{\text{synth}} \gets s \cdot \hat{\vxi}_T / \|\hat{\vxi}_T\|$
    \State Inverse PCA + destandardize: $\mathbf{x}_{\text{synth}} \gets \boldsymbol{\sigma} \odot (\mathbf{W}\,\mathbf{m}_{\text{synth}} + \bar{\mathbf{z}}) + \boldsymbol{\mu}$
    \State Split into visit records: $\mathbf{x}_{\text{synth}}^{(1)}, \mathbf{x}_{\text{synth}}^{(2)}, \mathbf{x}_{\text{synth}}^{(3)}$
\EndFor
\end{algorithmic}
\end{algorithm}

\subsection{Baseline and Validation Framework}
\label{sec:validation}

As a baseline, we generated synthetic patients by fitting a single multivariate
normal (MVN) distribution to the same $K{=}23$ concatenated 216-dimensional patient
profiles. Both methods operated on the same concatenated
representation. Each patient's three visits were joined into one vector before
generation, so that both methods had the opportunity to capture cross-visit
dependencies. For SA, this structure was preserved through the Hopfield energy
landscape operating on 18-dimensional PCA projections of the concatenated vectors.
For MVN, the $216 \times 216$ sample covariance matrix had rank at most~22 and was
regularized using Ledoit--Wolf shrinkage~\citep{Ledoit2004}, which inflated
eigenvalues in the 194 null dimensions.
To examine the role of the shared PCA representation, we also evaluated a fitted
two-component diagonal-covariance Gaussian mixture, a kernel-density estimator,
$k$-nearest-neighbor interpolation, and a Gaussian copula in the same 18-dimensional PCA
space. Their implementations and shared evaluation pipeline are described in Supplementary
Methods (PCA-space ablation).
We then evaluated both SA and MVN synthetic patients through four progressively stringent
validation levels.
\textbf{Level~1 (Marginal Plausibility)} tested whether individual feature
distributions matched, using per-feature means, standard deviations, and known
physiological relationships. \textbf{Level~2 (Joint Structure)} tested whether the
cross-visit covariance was preserved, by comparing full $216 \times 216$ correlation
matrices, eigenvalue spectra, and PCA projections between real, SA, and MVN
populations. \textbf{Level~3 (Conditional Structure)} tested whether rare subgroups
could be amplified while preserving condition-specific signatures, by generating
cohorts of 100 patients each from the Uncomplicated ($n{=}18$), PCOS ($n{=}3$), and
Developed~PE ($n{=}5$) subgroups using multiplicity-weighted sampling. \textbf{Level~4 (Mechanistic
Consistency)} tested whether synthetic patients produced biologically plausible
outputs under a separately specified, generator-blind mechanistic model calibrated
on the same real cohort. We used the Hockin--Mann BZ2012
coagulation model~\citep{Hockin2002,BrummelZiedins2012}, a system of 58 ODEs with 64 rate
constants, in which 5 rate constants were calibrated on Visit~1 real patients at the
population level and the remaining 59 were held at literature values. We ran the model
on all real and synthetic patients under two conditions (TF-only and TF+TM) and
compared predicted-versus-measured thrombin generation features. The primary metric
was cloud overlap: the fraction of synthetic predicted/measured ratios falling
within the 5th--95th percentile range of real ratios.

For the membership-inference analysis, we drew $40$ random partitions of the $23$ real
profiles. For each partition, we repeated the preprocessing using $15$ profiles, constructed
the memory matrix, and generated $100$ synthetic profiles; the other $8$ real profiles were
held out. We assigned each real profile a membership score equal to the negative Euclidean
distance to its nearest synthetic profile in standardized $216$-dimensional space, so a
higher score indicated stronger evidence that the profile had been stored. A receiver
operating characteristic curve compared the fraction of stored profiles correctly identified
with the fraction of held-out profiles incorrectly identified as the score threshold changed.
We calculated the area under this curve using the equivalent Mann--Whitney rank statistic,
with ties contributing one half, and reported the mean across the $40$ partitions. An AUC of
$0.5$ indicated chance discrimination, whereas $1.0$ indicated perfect discrimination.

\subsection{Robustness Tests}
\label{sec:robustness-benchmarks}

\begin{figure}[p]
\centering
\includegraphics[width=\textwidth]{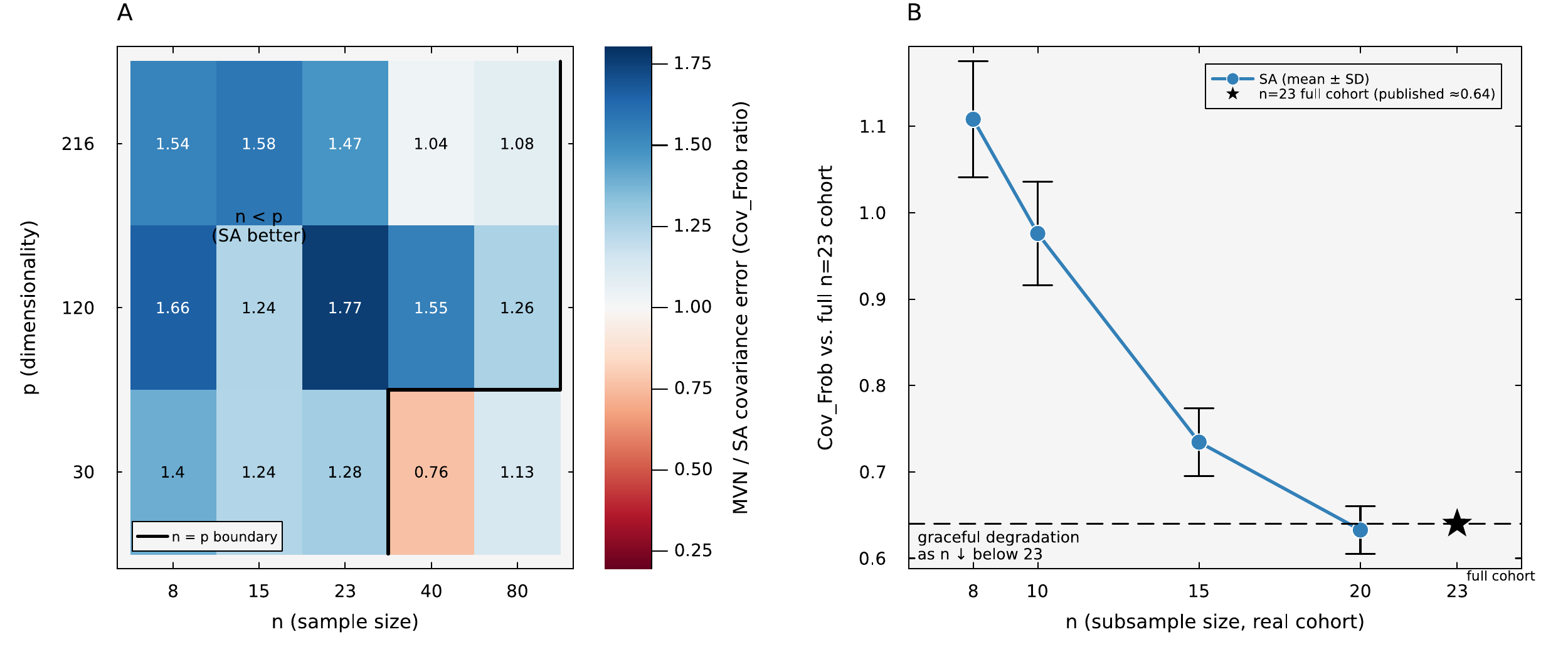}
\caption{\textbf{Population recovery and clinical-cohort subsampling.}
(A)~For each low-rank Gaussian setting, the stochastic-attention memory matrix was constructed
and a sample-covariance Gaussian generator was estimated from the same simulated training
cohort; both generators produced $100$ profiles. Color represented the ratio of their relative Frobenius errors against the known
population covariance (Gaussian error divided by stochastic-attention error), averaged over
latent ranks $3$, $5$, and~$10$. Values above one (blue) indicated lower error for stochastic
attention. Because the target covariance was fixed before the training cohort was drawn,
this measured population recovery rather than reproduction of the training data.
(B)~A stochastic-attention memory matrix was constructed from each of ten random subsets of
the clinical cohort at $n{=}20,15,10,$ and~$8$, and each matrix generated $100$ profiles.
Points and bars represented
the mean and standard deviation of the relative cross-visit correlation error against the
same full $23$-patient reference; the star represented generation using all $23$ patients.}
\label{fig:sim-recovery}
\end{figure}

Results from one clinical cohort could not show how stochastic-attention generation
behaved as training-set size, dimension, and nonlinear structure changed. We therefore used
three tests. The covariance test compared stochastic attention with a sample-covariance
Gaussian generator on low-rank Gaussian populations with known covariance. We generated one
training cohort for every tested combination from the factor model
$\mathbf{X}=\mathbf{Z}\mathbf{L}^{\top}+\boldsymbol{\epsilon}$, where
$\mathbf{Z}\in\mathbb{R}^{n\times r}$ and $\mathbf{L}\in\mathbb{R}^{p\times r}$ had
independent standard-normal entries and $\boldsymbol{\epsilon}$ had independent Gaussian
entries with variance $0.1$. This gave the known population covariance
$\boldsymbol{\Sigma}_{\mathrm{pop}}=\mathbf{L}\mathbf{L}^{\top}+0.1\mathbf{I}$. We tested
every combination of
$n\in\{8,15,23,40,80\}$, $p\in\{30,120,216\}$, and
$r\in\{3,5,10\}$. For each training cohort, we standardized the features, retained enough
principal components to explain $95\%$ of the variance, constructed the stochastic-attention
memory matrix from the resulting scores, and selected $\beta$ separately using the same
attention-entropy criterion as in the clinical analysis. Stochastic attention generated
$100$ profiles using ULA with $\alpha{=}0.01$, $T{=}2{,}000$, and empirical magnitude
rescaling. The Gaussian generator used the training mean and sample covariance. For numerical
sampling, diagonal jitter began at $10^{-8}\mathbf{I}$ and increased tenfold as needed through
$10^{-2}\mathbf{I}$; if Cholesky sampling still failed, we sampled in the nonnegative
eigenspace of the sample covariance. We applied no covariance shrinkage. The Gaussian
generator also produced $100$ profiles. Covariance error was
$\lVert\widehat{\boldsymbol{\Sigma}}_{\mathrm{gen}}-
\boldsymbol{\Sigma}_{\mathrm{pop}}\rVert_F/
\lVert\boldsymbol{\Sigma}_{\mathrm{pop}}\rVert_F$. Figure~\ref{fig:sim-recovery}A reports
Gaussian error divided by stochastic-attention error, averaged across the three latent ranks.
A ratio above one favored stochastic attention.

The remaining tests asked how stochastic attention changed with less clinical
information and how it compared with the Gaussian generator on a nonlinear population. For
the sample-size test, we drew ten subsets without replacement from the $23$ complete patients
at each $n\in\{20,15,10,8\}$. For every subset, we repeated the standardization, PCA, and
$\beta$ selection, constructed a new stochastic-attention memory matrix, generated $100$
profiles, and evaluated the generated cohort against the same $23$-patient reference. The
outcomes were relative cross-visit correlation error, pooled median marginal relative error,
and TF-only mechanistic cloud overlap. Figure~\ref{fig:sim-recovery}B shows the mean and
standard deviation of cross-visit error across the ten subsets. For the nonlinearity test, we
drew $t$ uniformly from $[-1,1]$ and defined an S-curve by $x_1=t$ and
$x_2=c\sin(\pi t)$. Viewed in the $(x_1,x_2)$ plane, this one-dimensional path resembled the
letter S because it had two opposing bends. Setting $c{=}0$ produced a straight line, and
increasing $c$ made the bends stronger. Each simulated profile was a noisy point near this
path. For each replicate, we used a new random transformation to spread the two coordinates
across $p{=}30$ or~$120$ simulated variables while preserving the curve's geometry. We then
added Gaussian noise with standard deviation $0.05$ in every direction. We used $n{=}23$ and curvature
$c\in\{0,0.25,0.5,1,1.5,2\}$, with five replicates per setting. From every training cohort, we
constructed a stochastic-attention memory matrix and estimated the same sample-covariance
Gaussian comparator; each generator produced $100$ profiles. We projected generated profiles
onto the known two-dimensional embedding plane and measured their mean distance from the
curve, thereby excluding noise orthogonal to that plane. We also measured relative covariance
error against the covariance of an independent $50{,}000$-profile population draw. For
the reported curvature threshold, we averaged stochastic-attention curve distance across the
five replicates and both dimensions at each curvature and identified the first nonzero value at
which this mean was at least twice its value for the straight-line population. All simulation
cells used deterministic, cell-specific random-number streams derived from seed~42.

\section*{Data Availability}
The code for synthetic patient generation, all validation scripts, and the
generated synthetic datasets are available at
\url{https://github.com/varnerlab/SA-generation-legacy-dataset-paper}.
The real patient data were collected under approval from the University of Vermont
Institutional Review Board; de-identified
data are available upon reasonable request to the corresponding author.

\section*{Acknowledgments}
This work was supported by NIH NHLBI R-33 HL~141787 (PIs Bernstein, Orfeo) and
NIH NHLBI R01 HL~71944 (PI Bernstein).

\section*{Author Contributions}
J.D.V.\ conceived the study, developed the SA pipeline and validation framework,
performed computational analyses, and wrote the manuscript. M.C.B.\ and T.O.\
oversaw the secondary analysis measurements of coagulation and fibrinolysis markers,
dynamic assays, and hormone measurements; provided domain expertise on coagulation
biology and thrombin generation assays; and reviewed the manuscript. C.M.\ was
involved in the enrollment of participants in the prospective research study and
collection of clinical information of the participants. I.B.\ designed and oversaw
the primary prospective research study of enrolled participants, provided clinical
interpretation, and reviewed the manuscript. All authors approved the final version.

\section*{Competing Interests}
The authors declare no competing interests.

\bibliographystyle{unsrtnat}
\bibliography{references}

\clearpage

\begin{table}[p]
\centering
\caption{\textbf{Demographics and clinical characteristics of the 23 patients
with complete longitudinal data across all three visits.} Values are reported as
mean~$\pm$~SD, median (range), or $n$~(\%) as appropriate.}
\label{tab:demographics}
\begin{tabular}{lccc}
\toprule
Characteristic & Mean $\pm$ SD & Median (Range) & $n$ (\%) \\
\midrule
Age at enrollment (yrs) & 30.2 $\pm$ 5.1 & 29 (20--41) & \\
\addlinespace
Race & & & \\
\quad White & & & 22 (96) \\
\quad Not disclosed & & & 1 (4) \\
\addlinespace
Prepregnancy BMI & 26.6 $\pm$ 5.3 & 25 (19.0--37.8) & \\
\addlinespace
Parity & & 0 (0--2) & \\
\quad Nulliparous & & & 16 (70) \\
\quad Parous & & & 7 (30) \\
\addlinespace
Enrollment condition & & & \\
\quad Healthy nulliparous & & & 14 (61) \\
\quad PCOS & & & 3 (13) \\
\quad Prior PE & & & 6 (26) \\
\addlinespace
Cycle day at V1 & 9.2 $\pm$ 3.6 & 10 (3--14) & \\
\addlinespace
Gestational age at V2 (days) & 89.0 $\pm$ 5.5 & & \\
Gestational age at V3 (days) & 217.0 $\pm$ 4.0 & & \\
\midrule
\multicolumn{4}{l}{\textit{Enrollment cohort $\times$ pregnancy outcome}} \\
\midrule
 & Uncomplicated & Developed PE & Total \\
\midrule
Healthy nulliparous & 12 & 2 & 14 \\
Prior PE & 4 & 2 & 6 \\
PCOS & 2 & 1 & 3 \\
\midrule
Total & 18 & 5 & 23 \\
\bottomrule
\end{tabular}
\end{table}

\begin{table}[p]
\centering
\caption{Per-visit mean relative error (MRE) between real and SA-generated synthetic
patients for representative coagulation features, computed as
$|\bar{x}_{\text{synth}} - \bar{x}_{\text{real}}| / |\bar{x}_{\text{real}}|$.}
\label{tab:summary}
\begin{tabular}{lccc}
\toprule
Feature & Visit 1 & Visit 2 & Visit 3 \\
\midrule
Factor II (\%) & 0.018 & 0.017 & 0.007 \\
Factor VIII (\%) & 0.018 & 0.008 & 0.005 \\
AT (\%) & 0.025 & 0.001 & 0.008 \\
Fibrinogen (mg/dL) & 0.005 & 0.009 & $<$0.001 \\
TF Peak (nM) & 0.025 & 0.002 & 0.017 \\
TF ETP (nM$\cdot$min) & 0.013 & 0.008 & 0.023 \\
\bottomrule
\end{tabular}
\end{table}

\clearpage


\clearpage

\clearpage

\clearpage

\clearpage

\clearpage

\clearpage

\clearpage

\setcounter{section}{0}
\setcounter{figure}{0}
\setcounter{table}{0}
\setcounter{equation}{0}
\renewcommand{\thesection}{S\arabic{section}}
\renewcommand{\thefigure}{S\arabic{figure}}
\renewcommand{\thetable}{S\arabic{table}}
\renewcommand{\theequation}{S\arabic{equation}}

\section*{Supplementary Information}

\subsection*{Supplementary Methods}

\subsubsection*{PCA-space ablation}

We compared the canonical stochastic-attention cohort with four baseline generators
operating on the same unnormalized 18-dimensional PCA coordinates. Each baseline generated
$100$ profiles with a separate random-number stream initialized using seed~42. The fitted
two-component Gaussian-mixture baseline used diagonal covariance matrices, $100$
expectation--maximization iterations, and a variance floor of $10^{-3}$. The kernel-density sampler chose
a stored PCA point uniformly and added Gaussian noise with the empirical PCA covariance and
Scott bandwidth $h=K^{-1/(d+4)}$; $10^{-8}\mathbf{I}$ was added for numerical factorization.
The nearest-neighbor sampler chose a stored point, selected one of its three nearest
neighbors uniformly, and interpolated between them with a coefficient drawn from
$\operatorname{Uniform}(0,1)$. The Gaussian copula transformed empirical marginal ranks to
normal scores, estimated their correlation matrix with $10^{-6}\mathbf{I}$ added for
factorization, and mapped sampled scores back through the empirical marginal quantiles. All
samples were passed through the same inverse-PCA and destandardization pipeline and evaluated
with the same marginal, cross-visit, mechanistic, and novelty calculations.

\subsubsection*{Selecting the inverse-temperature operating point}

For each candidate $\beta$, we computed the Shannon entropy of
$\text{softmax}(\beta\,\mhat{X}^\top\vxi)$ at $20$ stored-pattern probes, averaged the values,
and divided by $\log K$. Zero means one pattern carries all attention; one means all patterns
have equal attention. We evaluated $80$ logarithmically spaced values from $0.1$ to $1000$ and
selected $\beta^*$ at the most negative finite-difference second derivative with respect to
$\log\beta$. This marked the shift from shared to concentrated attention. For conditioned
sampling, we repeated the calculation with the $\log r_k$ bias, producing
$\beta^*(\rho)$. The random-pattern scale $\sqrt{d}$ was the theoretical reference.

\subsection*{Supplementary Tables}

\begin{table}[ht]
\centering
\caption{\textbf{Inverse-temperature sensitivity.} Generation metrics across
$\beta/\beta^*$, where $\beta^*\approx2.94$ was selected from the attention-entropy
transition in the 18-dimensional PCA space. The bold row marks the reported value.
Med MRE is the median relative error of feature means; novelty is
$1-\max_k\cos(\hat{\xi},\hat{m}_k)$.}
\label{stab:beta-sweep}
\begin{tabular}{rrcccc}
\toprule
$\beta/\beta^*$ & $\beta$ & Noise scale & PC1 std ratio & Novelty & Med MRE (\%) \\
\midrule
0.1 & 0.29 & 0.261 & 0.538 & 0.542 & 0.7 \\
0.3 & 0.88 & 0.151 & 0.550 & 0.536 & 0.7 \\
0.5 & 1.47 & 0.117 & 0.562 & 0.528 & 0.7 \\
\textbf{1.0} & \textbf{2.94} & \textbf{0.082} & \textbf{0.589} & \textbf{0.502} & \textbf{0.6} \\
1.5 & 4.41 & 0.067 & 0.613 & 0.474 & 0.7 \\
2.0 & 5.88 & 0.058 & 0.638 & 0.446 & 0.7 \\
3.0 & 8.82 & 0.048 & 0.689 & 0.403 & 0.8 \\
\bottomrule
\end{tabular}
\end{table}

\begin{table}[ht]
\centering
\caption{\textbf{Multiplicity weighting parameters for conditional generation.}
For the unit-normalized memories used here, the multiplicity ratio $\rho$ was computed
to place a target fraction $f_{\text{target}} = 0.80$ of finite-mixture component
probability on the designated patterns. $K_{\text{eff}}$ is the participation ratio of
those component probabilities and describes their concentration across repeated draws.}
\label{stab:multiplicity}
\begin{tabular}{lccccc}
\toprule
Condition & $K_{\text{sub}}$ & $K_{\text{bg}}$ & $\rho$ & $f_{\text{eff}}$ & $K_{\text{eff}}$ \\
\midrule
Uncomplicated  & 18 & 5  & 1.11  & 0.80 & 23.0 \\
PCOS           & 3  & 20 & 26.67 & 0.80 & 4.6 \\
Developed PE   & 5  & 18 & 14.40 & 0.80 & 7.7 \\
\bottomrule
\end{tabular}
\end{table}

\begin{table}[ht]
\centering
\caption{\textbf{SA generation hyperparameters.} The same values were used for
unconditioned and conditioned generation except that $\beta^*$ was recomputed with the
multiplicity bias for conditioned cohorts.}
\label{stab:hyperparams}
\begin{tabular}{lll}
\toprule
Parameter & Value & Description \\
\midrule
$\alpha$ & 0.01 & Langevin step size (see Table~\ref{tab:ula-vs-mala}) \\
$T$ & 2,000 & Langevin iterations per sample \\
$\beta^*$ & 2.94 & Entropy-transition point (unconditioned) \\
PCA threshold & 95\% & Variance retained \\
$d_{\text{PCA}}$ & 18 & PCA dimensionality \\
$N$ & 100 & Synthetic patients per cohort \\
Random seed & 42 & For reproducibility \\
$f_{\text{target}}$ & 0.80 & Designated-component probability (conditioned) \\
\bottomrule
\end{tabular}
\end{table}

\begin{table}[ht]
\centering
\caption{\textbf{BZ2012 ODE model calibration.} Five rate constants were
calibrated on Visit~1 real patients ($n{=}23$) using bounded Nelder--Mead
optimization in log-scale-factor space. The cost function was the mean
normalized relative squared error across five TGA features under TF-only
conditions. Bounds: $\pm\log(100)$. Convergence: $f_{\text{tol}} = 10^{-6}$,
500 iterations, 11 random restarts.}
\label{stab:ode-calibration}
\begin{tabular}{llrrl}
\toprule
Rate constant & Role & Default & Calibrated & Scale \\
\midrule
prothrombinase $k_{\text{cat}}$ & Thrombin burst & 63.5 s$^{-1}$ & 21.94 s$^{-1}$ & 0.35$\times$ \\
intrinsic Xase $k_{\text{cat}}$ & Amplification Xa & 8.2 s$^{-1}$ & 0.169 s$^{-1}$ & 0.021$\times$ \\
extrinsic Xase $k_{\text{cat}}$ & Initiation Xa & 6.0 s$^{-1}$ & 93.2 s$^{-1}$ & 15.5$\times$ \\
PC activation $k_{\text{cat}}$ & Protein C activation & 0.41 s$^{-1}$ & 0.890 s$^{-1}$ & 2.17$\times$ \\
mIIa conversion $k$ & mIIa$\to$IIa & $2.3{\times}10^8$ & $1.15{\times}10^9$ & 5.01$\times$ \\
\bottomrule
\multicolumn{5}{l}{\small Units for mIIa conversion: M$^{-1}$s$^{-1}$} \\
\end{tabular}
\end{table}

\begin{table}[ht]
\centering
\caption{\textbf{Summary of mean relative errors across all 72 features.}
Per-visit distribution of MREs between real and SA-generated synthetic patient
means. The full feature-by-feature table (72 features $\times$ 3 visits = 216
entries) is provided in the code repository as
\texttt{paper\_summary\_statistics.csv}.}
\label{stab:mre-summary}
{\small
\setlength{\tabcolsep}{4pt}
\begin{tabular}{lrrrrrr}
\toprule
Visit & Median & Mean & 25th pct. & 75th pct. & Maximum & MRE below 5\% \\
\midrule
V1 (baseline)   & 0.022 & 0.029 & 0.008 & 0.039 & 0.158 & 59/72 (81.9\%) \\
V2 (1st tri)    & 0.009 & 0.021 & 0.004 & 0.022 & 0.224 & 65/72 (90.3\%) \\
V3 (3rd tri)    & 0.011 & 0.016 & 0.005 & 0.024 & 0.062 & 69/72 (95.8\%) \\
\midrule
\textbf{Pooled (all 3 visits)} & \textbf{0.012} & \textbf{0.022} & \textbf{0.005} & \textbf{0.028} & \textbf{0.224} & \textbf{193/216 (89.4\%)} \\
\bottomrule
\multicolumn{7}{p{0.95\textwidth}}{\small Pooled median MRE 95\% bootstrap CI: (0.98\%, 1.60\%) over 10{,}000 resamples (seed~42).}
\end{tabular}
}
\end{table}

\begin{table}[ht]
\centering
\caption{\textbf{Comparison with deep generative baselines.} CTGAN and TVAE
were trained on the same 90 per-visit records (23 patients $\times$ 3 visits,
72 features) at three epoch counts. Results for SA and MVN are shown for reference.
The CTGAN model had median MRE near 19\%, consistent with mode collapse at this sample size.
The TVAE model matched marginal fidelity at high epoch counts but could not capture
cross-visit covariance or perform conditional generation.}
\label{stab:ctgan}
\begin{tabular}{lcccc}
\toprule
Method & Epochs & Med MRE & Med KS & Corr MAE \\
\midrule
SA              & --   & 0.019 & 0.169 & 0.200 \\
MVN             & --   & 0.024 & 0.161 & 0.090 \\
\midrule
CTGAN           & 300   & 0.189 & 0.364 & 0.261 \\
CTGAN           & 1,000 & 0.195 & 0.349 & 0.265 \\
CTGAN           & 3,000 & 0.187 & 0.341 & 0.180 \\
\midrule
TVAE            & 300   & 0.037 & 0.174 & 0.152 \\
TVAE            & 1,000 & 0.026 & 0.248 & 0.082 \\
TVAE            & 3,000 & 0.018 & 0.224 & 0.092 \\
\bottomrule
\end{tabular}
\end{table}

\begin{table}[ht]
\centering
\caption{\textbf{PCA variance threshold sensitivity analysis.} SA generation
quality is stable across thresholds from 85\% to 99\%. Median MRE remains
between 1.0\% and 1.7\% across all thresholds. The 95\% threshold used in the
main text (bold) is not a critical design choice.}
\label{stab:pca-sensitivity}
\begin{tabular}{rrccccc}
\toprule
Threshold & $d_{\text{PCA}}$ & $K/d$ & $\beta^*$ & Novelty & Med MRE & Corr MAE \\
\midrule
85\% & 13 & 1.77 & 2.94 & 0.420 & 0.0098 & 0.124 \\
90\% & 15 & 1.53 & 2.94 & 0.477 & 0.0160 & 0.126 \\
\textbf{95\%} & \textbf{18} & \textbf{1.28} & \textbf{2.94} & \textbf{0.500} & \textbf{0.0124} & \textbf{0.143} \\
97.5\% & 20 & 1.15 & 2.94 & 0.543 & 0.0120 & 0.135 \\
99\% & 21 & 1.10 & 2.94 & 0.541 & 0.0172 & 0.138 \\
\bottomrule
\end{tabular}
\end{table}

\begin{table}[ht]
\centering
\caption{\textbf{ULA and MALA sampling diagnostics.} Ten chains of 5{,}000
iterations each were run with both the Unadjusted Langevin Algorithm (ULA)
and the Metropolis-Adjusted Langevin Algorithm (MALA) at step size
$\alpha = 0.01$ and $\beta^* = 2.94$ on the 18-dimensional PCA memory
($K{=}23$ patterns). The MALA acceptance rate was $99.9\%$, and the energy and effective
sample-size estimates were similar between methods, indicating little discretization bias.
This comparison bounded the
discretization error of the reported sampler. The direct reference for the
target itself was the analytic sampler of Eq.~\maineqref{eq:exact_mixture},
introduced below.}
\label{tab:ula-vs-mala}
\begin{tabular}{lcc}
\toprule
Metric & ULA & MALA \\
\midrule
Acceptance rate (\%)         & -- (no rejection) & $99.9 \pm 0.03$ \\
Mean energy (post-burn-in)   & $1.932 \pm 0.141$ & $1.886 \pm 0.231$ \\
$\tau_{\text{int}}$          & $109.5 \pm 49.5$  & $106.7 \pm 30.4$ \\
Effective sample size        & $41.8 \pm 14.2$   & $40.0 \pm 10.2$ \\
\bottomrule
\multicolumn{3}{p{0.85\textwidth}}{\small Burn-in: 1{,}000 iterations discarded. $\tau_{\text{int}}$: integrated
autocorrelation time estimated with a windowed estimator (threshold 0.05).
ESS = $(T - T_{\text{burn}}) / \tau_{\text{int}}$.}
\end{tabular}
\end{table}

\begin{table}[ht]
\centering
\caption{\textbf{Proximity to stored profiles.} Two measures compared synthetic and stored
profiles. Novelty is the angular distance between a synthetic profile and its
nearest stored pattern at $\beta = \beta^*$. Nearest-neighbor distances are
computed in the per-visit standardized 72-dimensional feature space.}
\label{stab:memorization}
\begin{tabular}{lc}
\toprule
Metric & Value \\
\midrule
Mean novelty score, $1 - \max_k \cos(\hat{\xi}, \hat{m}_k)$ & 0.50 \\
Median synth--real nearest-neighbor distance (std.\ units) & 6.14 \\
Median real--real nearest-neighbor distance (std.\ units)  & 6.87 \\
Distance ratio (synth--real / real--real)                  & 0.89 \\
\bottomrule
\end{tabular}
\end{table}

\begin{table}[ht]
\centering
\caption{\textbf{Descriptive bootstrap Mann--Whitney diagnostic per feature and
condition.} For each of 24 feature--condition pairs, we subsampled the
synthetic cohort to match the real sample size, applied a Mann--Whitney U test,
and repeated 1{,}000 times. Frac.\ $p > 0.05$ is the fraction of replicates in
which the test detected no difference at the available sample size. This
power-limited diagnostic is not an equivalence test.
Twenty of 24 pairs had $p > 0.05$ in $\geq 90\%$ of replicates; the four
below-threshold pairs (italicized) involved high-variance features in the smallest
subgroups.}
\label{stab:bootstrap-mw}
\begin{tabular}{llcc}
\toprule
Condition & Feature & MRE & Frac.\ $p > 0.05$ \\
\midrule
Uncomplicated ($n{=}18$) & FVIII   & 0.057 & 0.952 \\
Uncomplicated ($n{=}18$) & vWF     & 0.022 & 0.999 \\
Uncomplicated ($n{=}18$) & FIX     & 0.018 & 0.998 \\
Uncomplicated ($n{=}18$) & FV      & 0.006 & 0.996 \\
Uncomplicated ($n{=}18$) & $\alpha_2$-AP & 0.019 & 1.000 \\
Uncomplicated ($n{=}18$) & FVII    & 0.057 & 0.941 \\
Uncomplicated ($n{=}18$) & Fbgn    & 0.017 & 0.987 \\
Uncomplicated ($n{=}18$) & Plgn    & 0.020 & 0.988 \\
\midrule
PCOS ($n{=}3$)  & FVIII   & 0.109 & 0.985 \\
PCOS ($n{=}3$)  & vWF     & 0.150 & 0.959 \\
PCOS ($n{=}3$)  & \textit{FIX}     & \textit{0.121} & \textit{0.873} \\
PCOS ($n{=}3$)  & FV      & 0.057 & 0.944 \\
PCOS ($n{=}3$)  & $\alpha_2$-AP & 0.051 & 0.987 \\
PCOS ($n{=}3$)  & FVII    & 0.029 & 0.999 \\
PCOS ($n{=}3$)  & Fbgn    & 0.043 & 0.992 \\
PCOS ($n{=}3$)  & \textit{Plgn}    & \textit{0.153} & \textit{0.615} \\
\midrule
Developed PE ($n{=}5$) & FVIII   & 0.097 & 0.932 \\
Developed PE ($n{=}5$) & vWF     & 0.010 & 0.986 \\
Developed PE ($n{=}5$) & \textit{FIX}     & \textit{0.108} & \textit{0.780} \\
Developed PE ($n{=}5$) & FV      & 0.006 & 0.987 \\
Developed PE ($n{=}5$) & $\alpha_2$-AP & 0.062 & 0.950 \\
Developed PE ($n{=}5$) & FVII    & 0.045 & 0.993 \\
Developed PE ($n{=}5$) & Fbgn    & 0.048 & 0.996 \\
Developed PE ($n{=}5$) & \textit{Plgn}    & \textit{0.093} & \textit{0.814} \\
\bottomrule
\multicolumn{4}{p{0.95\textwidth}}{\small Median Frac.\ $p > 0.05$ across all 24 pairs: 0.986. Failing
pairs ($< 0.90$, italicized): PCOS--FIX, PCOS--Plgn, DPE--FIX, DPE--Plgn.}
\end{tabular}
\end{table}

\begin{table}[ht]
\centering
\caption{\textbf{Mechanistic validation diagnostics per feature and
condition.} Cloud overlap is the fraction of synthetic ODE-predicted/measured
ratios falling within the 5th--95th percentile of the corresponding real
distribution. KS $D$ and KS $p$ are from a two-sample Kolmogorov--Smirnov test
on the same ratio distributions. Real $n{=}69$ (23 patients $\times$ 3 visits);
synthetic $n{=}300$ (100 patients $\times$ 3 visits).}
\label{stab:mechanistic-validation}
\begin{tabular}{llccc}
\toprule
Condition & TGA Feature & Cloud overlap & KS $D$ & KS $p$ \\
\midrule
TF-only & Lagtime          & 0.92 & 0.112 & 0.48 \\
TF-only & Peak             & 0.91 & 0.081 & 0.86 \\
TF-only & T.Peak           & 0.92 & 0.102 & 0.61 \\
TF-only & Max Rate         & 0.91 & 0.083 & 0.84 \\
TF-only & ETP              & 0.83 & 0.123 & 0.36 \\
\midrule
TF+TM   & Lagtime          & 0.92 & 0.127 & 0.32 \\
TF+TM   & Peak             & 0.88 & 0.079 & 0.88 \\
TF+TM   & T.Peak           & 0.91 & 0.110 & 0.51 \\
TF+TM   & Max Rate         & 0.88 & 0.096 & 0.67 \\
TF+TM   & ETP              & 0.89 & 0.112 & 0.48 \\
\bottomrule
\multicolumn{5}{p{0.95\textwidth}}{\small Under TF-only conditions, the cloud overlap
range was 0.83--0.92. Under TF+TM conditions, the range was 0.88--0.92. Pooled across both
conditions and all five features, the overlap was 0.902.
KS $D$ range across both conditions: 0.079--0.127, all $p > 0.30$.}
\end{tabular}
\end{table}

\begin{table}[ht]
\centering
\caption{\textbf{Downstream utility: per-feature held-out V2/V3 performance.}
Median relative error of the BZ2012 ODE-predicted TGA values against the held-out
real V2 and V3 measurements, for the model calibrated on real V1 patients
($K{=}23$) and the model calibrated on synthetic V1 patients ($N{=}100$). The
synth-calibrated model achieves slightly lower error on every feature. Synth/Real
ratio $< 1$ favors the synth-calibrated model.}
\label{stab:downstream-utility}
\begin{tabular}{lccc}
\toprule
TGA Feature & Real-cal MRE & Synth-cal MRE & Synth/Real \\
\midrule
Lagtime           & 0.165 & 0.162 & 0.98$\times$ \\
Peak              & 0.097 & 0.087 & 0.90$\times$ \\
T.Peak            & 0.330 & 0.306 & 0.93$\times$ \\
Max Rate          & 0.286 & 0.259 & 0.91$\times$ \\
ETP               & 0.197 & 0.183 & 0.93$\times$ \\
\midrule
\textbf{Overall median} & \textbf{0.201} & \textbf{0.189} & \textbf{0.94$\times$} \\
\bottomrule
\end{tabular}
\end{table}

\begin{table}[ht]
\centering
\caption{\textbf{PCA-space ablation.} Five generators used the same $d{=}18$ PCA
subspace and evaluation. PCA+GMM is the fitted two-component diagonal-covariance baseline,
not the exact stochastic-attention mixture. MRE measures marginal fidelity. Mechanistic overlap is the fraction
of pooled synthetic predicted-to-recorded TGA ratios within the real 5th--95th percentile
range. Median novelty is $1-\max_k\cos(\hat{\vxi},\hat{\mathbf{m}}_k)$; Frac.\ novel is the
fraction above $0.2$. Cross-visit Frobenius error measures deviation from the real correlation
matrix. Lower error indicates closer in-sample reproduction and can reward memorization.}
\label{stab:pca-ablation}
\begin{tabular}{lccccc}
\toprule
Method & MRE (\%) & Cross-visit Frob. & Mech.\ overlap & Median novelty & Frac.\ novel \\
\midrule
SA         & 1.24 & 0.641 & 0.902 & 0.514 & 1.00 \\
PCA+GMM    & 1.13 & 0.422 & 0.918 & 0.497 & 0.94 \\
PCA+KDE    & 1.94 & 0.337 & 0.823 & 0.249 & 0.65 \\
PCA+kNN    & 1.63 & 0.587 & 0.938 & 0.051 & 0.05 \\
PCA+Copula & 1.34 & 0.417 & 0.934 & 0.462 & 1.00 \\
\bottomrule
\end{tabular}
\end{table}

\begin{table}[ht]
\centering
\caption{\textbf{Convex-hull geometry and plausibility at $K{=}23$ in $d{=}18$.}
We compared $100$ synthetic profiles with leave-one-out real profiles using hulls of $22$
real points. For each synthetic profile, we omitted its nearest real profile. The radial
overshoot $1/t^{*}$ used the point where a ray from the hull centroid exited the hull.
All profiles lay outside their comparison hulls. Median hull distance was $11.0$ for synthetic
profiles and $11.9$ for real profiles ($p{=}0.07$); radial overshoot was smaller for synthetic
profiles ($p<10^{-4}$). These descriptive tests treated the synthetic profiles as independent,
so their significance was optimistic. Hull membership had little resolution because every
real profile formed an outer corner, and a unit-normalized generated direction could remain
inside only by matching a stored direction.
The biological check required nonnegative values and limited coagulation-factor and TGA values
to five real-cohort standard deviations from the mean. Fifty-four synthetic profiles passed.
The other $46$ contained negative estradiol or progesterone values: estradiol was negative in
$19$ profiles and progesterone in $37$, with $10$ overlapping; the minima were $-2109.8$ and
$-75.8$. No variable was constrained to be positive, and linear decoding had unbounded support.
Hormones were closer to zero relative to their observed variation than the coagulation factors,
which explained why only hormones crossed zero in these draws but did not guarantee positive
factors in other draws. All nine factor inputs to BZ2012 were positive, so all $100$ profiles
entered the mechanistic check. The unconstrained MVN baseline produced negative hormones in
$43$ profiles, compared with $46$ for stochastic attention.}
\label{stab:hull-geometry}
{\small
\setlength{\tabcolsep}{5pt}
\begin{tabular}{lcc}
\toprule
Quantity & Synthetic ($n{=}100$) & Real, leave-one-out ($n{=}23$) \\
\midrule
Fraction inside hull                        & $0.00$ & $0.00$ \\
Median distance to hull                     & $11.0$ & $11.9$ \\
Median radial overshoot $1/t^{*}$           & $10.5$ & $15.2$ \\
\midrule
\multicolumn{3}{l}{\emph{Hull geometry of the real cohort}} \\
Median patient radius                       & \multicolumn{2}{c}{$13.8$} \\
Median hull extent, generic direction       & \multicolumn{2}{c}{$2.0$} \\
Largest $t^{*}$ attained on the unit sphere & \multicolumn{2}{c}{$0.231$} \\
\midrule
\multicolumn{3}{l}{\emph{Plausibility of the out-of-hull synthetic patients}} \\
Biological constraints satisfied            & \multicolumn{2}{c}{$54/100$} \\
Mechanistic band, per feature               & \multicolumn{2}{c}{$83$ to $92\%$} \\
Mechanistic band, all features and visits   & \multicolumn{2}{c}{$41/100$} \\
Both biological and strict mechanistic criteria & \multicolumn{2}{c}{$24/100$} \\
\bottomrule
\end{tabular}
}
\end{table}

\begin{table}[ht]
\centering
\caption{\textbf{Membership inference recovered memory membership, not unseen
patients.} Median nearest-neighbor distances in the standardized $216$-dimensional space,
and the test's discrimination. For each split, we constructed the memory matrix from $15$
patients and scored each real record by its distance to the resulting synthetic set,
treating the $15$ included patients as members and the $8$ held-out patients as
non-members. Every quantity came from the same protocol. We drew $40$ partitions at random
with the generation seed held fixed, so that only the partition varied, and pooled the
distances across all partitions. Non-members sat at the same distance from the synthetic
cohort as unrelated real patients sat from one another. The synthetic cohort
carried no proximity signal about patients the generator never saw, and the discrimination
came entirely from members sitting closer. The area under the receiver operating
characteristic curve (AUC), calculated using the equivalent Mann--Whitney rank statistic,
measured how often the test assigned stronger membership evidence to a randomly chosen member
than to a randomly chosen non-member; $0.5$ indicated no discrimination and $1.0$ indicated
perfect discrimination.}
\label{stab:membership}
\begin{tabular}{lc}
\toprule
Quantity & Value \\
\midrule
\multicolumn{2}{l}{\emph{Median nearest-neighbor distance}} \\
Memory-matrix member to nearest synthetic      & $10.7$ \\
Held-out non-member to nearest synthetic       & $15.4$ \\
Real to nearest real, leave-one-out            & $15.9$ \\
Synthetic to nearest real                      & $13.8$ \\
\midrule
\multicolumn{2}{l}{\emph{Membership-inference AUC over $40$ random splits}} \\
Mean                                           & $0.97$ \\
Standard deviation                             & $0.02$ \\
Median                                         & $0.98$ \\
Minimum                                        & $0.90$ \\
Maximum                                        & $1.00$ \\
Splits reaching perfect separation             & $9/40$ \\
Splits with AUC above $0.9$                    & $39/40$ \\
\bottomrule
\end{tabular}
\end{table}

\FloatBarrier

\subsection*{Supplementary Figures}

\begin{figure}[ht]
\centering
\includegraphics[width=\textwidth]{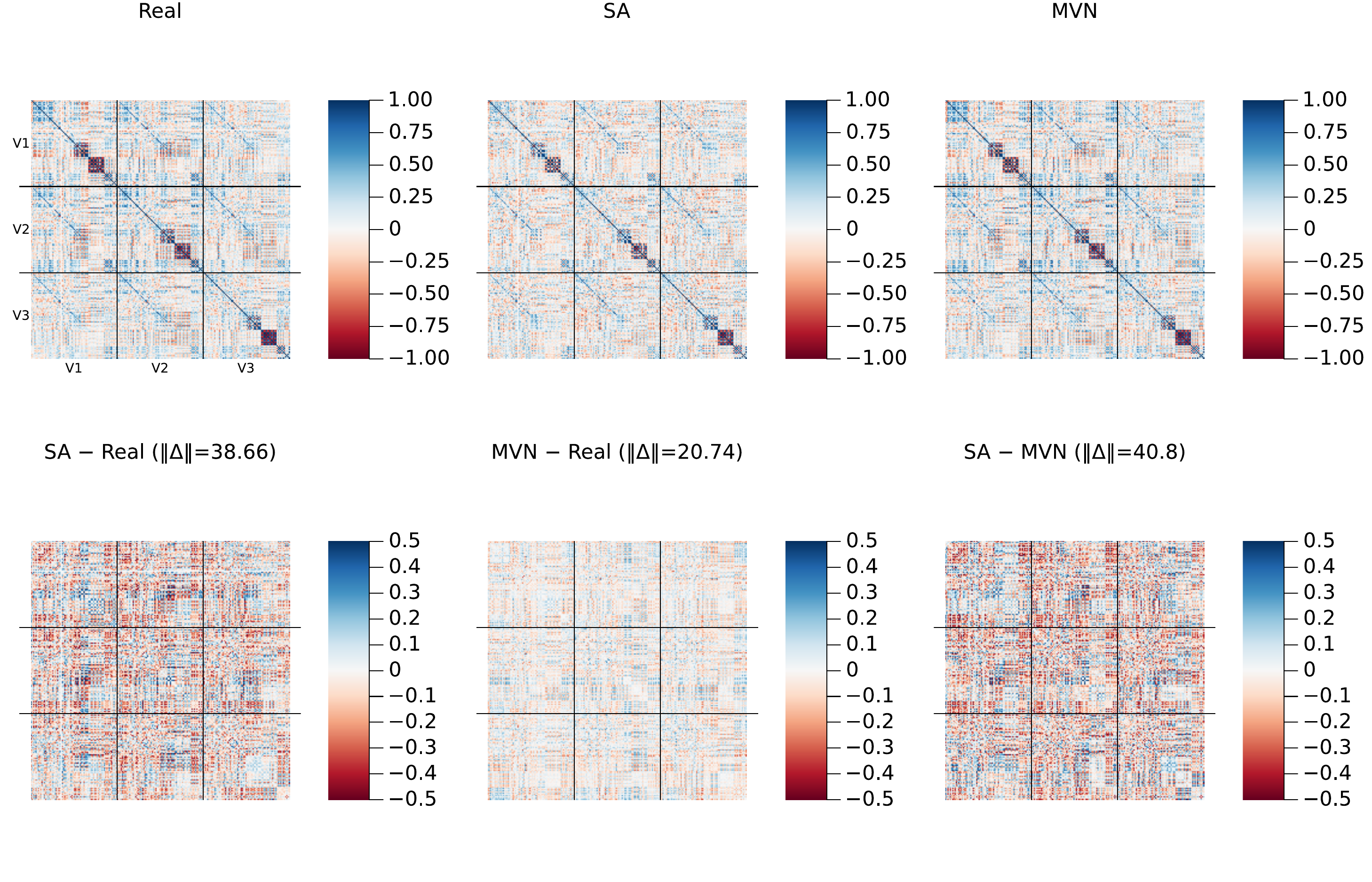}
\caption{\textbf{Cross-visit correlation structure (amplified residual scale).}
Same layout as the main-text correlation figure, but with the bottom-row
residual panels plotted on a $\pm 0.5$ color scale (vs.\ $\pm 1.0$ in the main
text) to reveal fine structure. Frobenius norms of each residual matrix are
shown in the panel titles.}
\label{sfig:cross-visit-corr-supp}
\end{figure}

\begin{figure}[ht]
\centering
\includegraphics[width=0.75\textwidth]{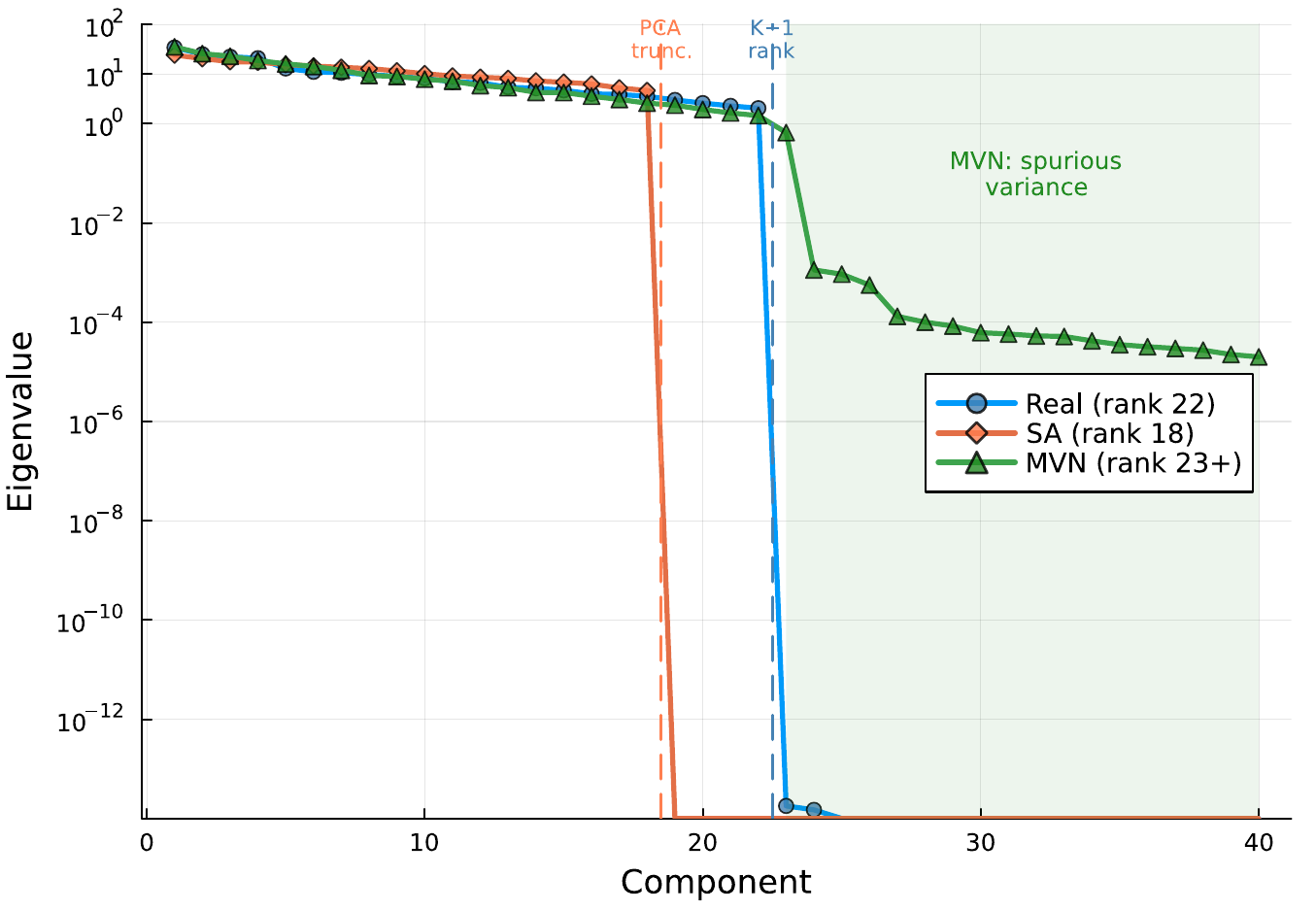}
\caption{\textbf{Covariance eigenvalue spectrum.} Log-scale eigenvalues of the
$216 \times 216$ sample covariance for real ($K{=}23$, rank~22), SA ($N{=}100$,
rank~18), and MVN ($N{=}100$) populations. Dashed vertical lines mark the SA
PCA truncation point (component~18) and the data rank limit (component~22).
Ledoit--Wolf regularization for MVN inflates eigenvalues beyond component~22
(green shaded region), introducing spurious variance in 194 dimensions where the
data contain no signal.}
\label{sfig:eigenvalue-spectrum}
\end{figure}

\begin{figure}[ht]
\centering
\includegraphics[width=0.65\textwidth]{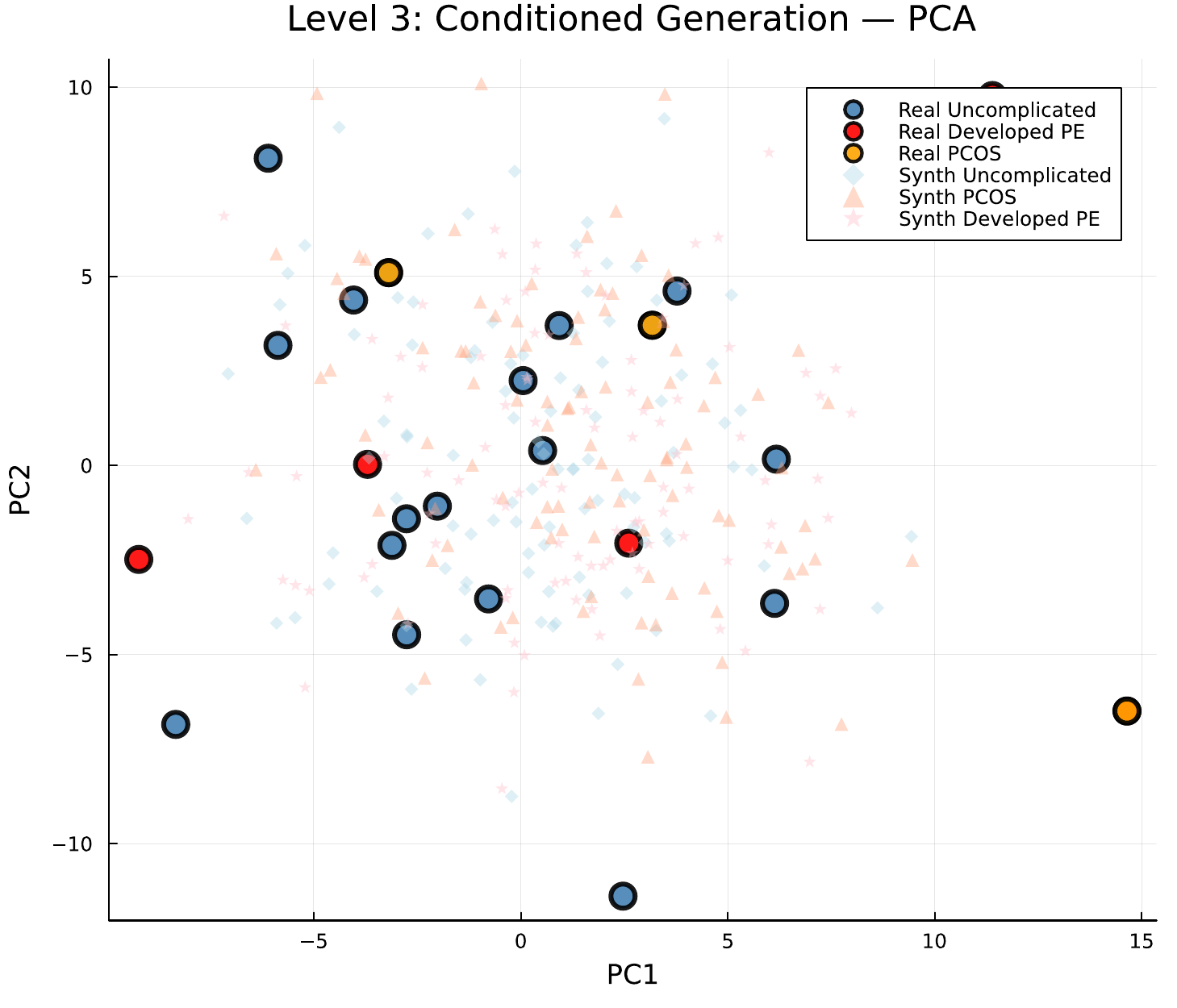}
\caption{\textbf{Conditioned generation: PCA projection (pooled).} Large
markers: real patients (blue = Uncomplicated, $n{=}18$; orange = PCOS, $n{=}3$;
red = Developed~PE, $n{=}5$). Small markers: synthetic patients ($N{=}100$ per
condition). With only 3 PCOS and 5 Developed~PE patients, the subgroups do not form
clearly separable clusters in the first two principal components, but the
synthetic cohorts concentrate around their respective real counterparts.}
\label{sfig:conditioned-pca}
\end{figure}

\begin{figure}[ht]
\centering
\includegraphics[width=\textwidth]{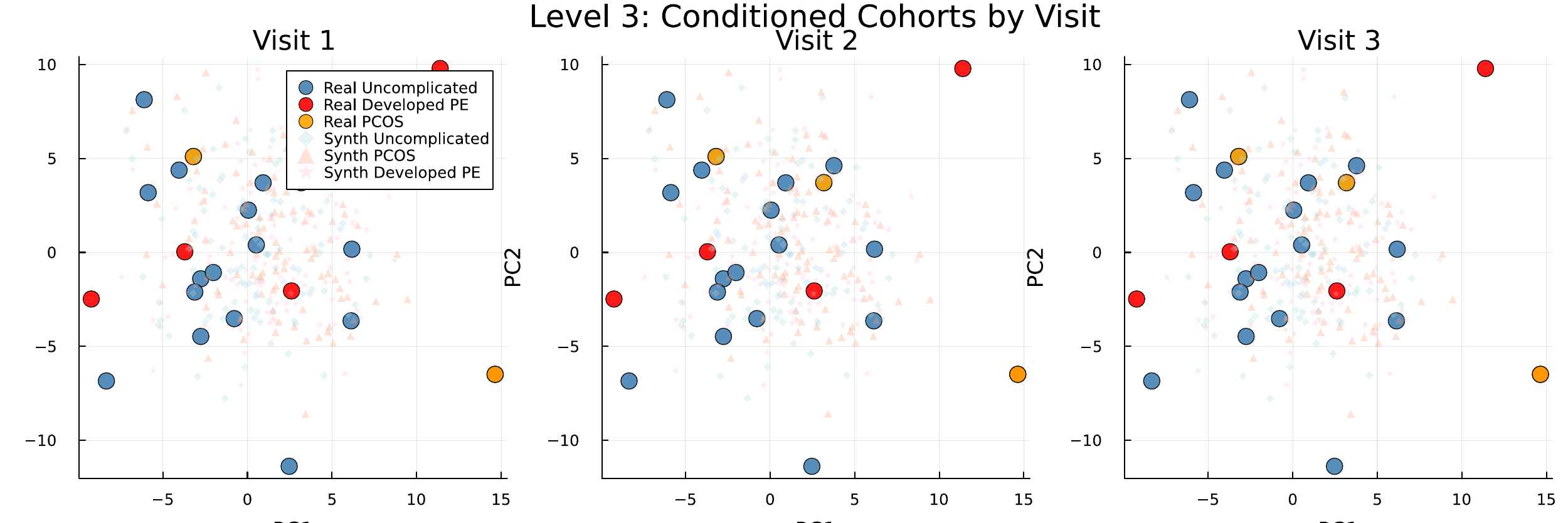}
\caption{\textbf{Conditioned generation: PCA projection by visit.} Same as
Fig.~\ref{sfig:conditioned-pca} but separated by visit (V1, V2,
V3). Condition-specific clustering is maintained within each individual visit.}
\label{sfig:conditioned-by-visit}
\end{figure}

\begin{figure}[ht]
\centering
\includegraphics[width=\textwidth]{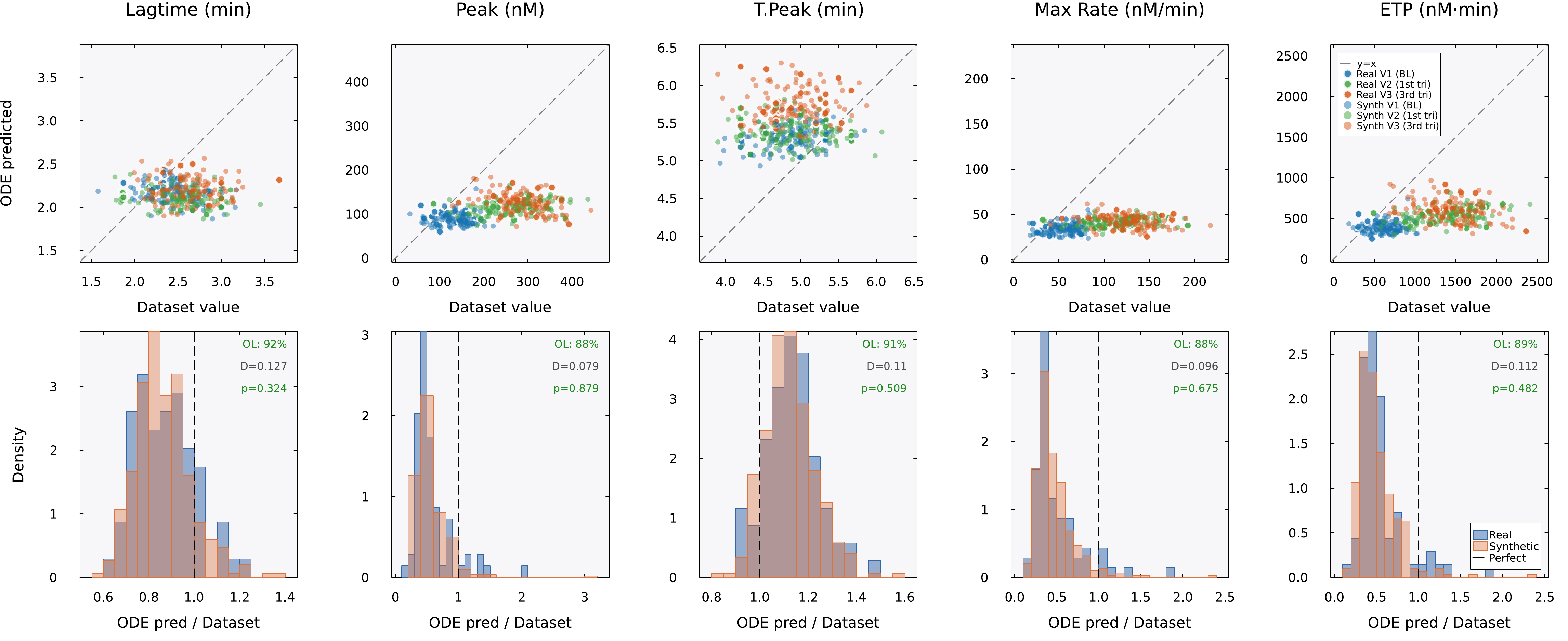}
\caption{\textbf{Mechanistic validation (TF+TM condition).} Same layout as
main-text Fig.~6 but under TF+TM conditions. The model systematically
underpredicts peak and maximum rate ($\approx 0.5\times$ measured) for both
real and synthetic patients, reflecting overcorrection of the protein~C
anticoagulant pathway. Cloud overlap was 88--92\%, showing similar model behavior for
real and synthetic profiles despite imperfect calibration.}
\label{sfig:mechanistic-tm-combined}
\end{figure}

\begin{figure}[ht]
\centering
\includegraphics[width=\textwidth]{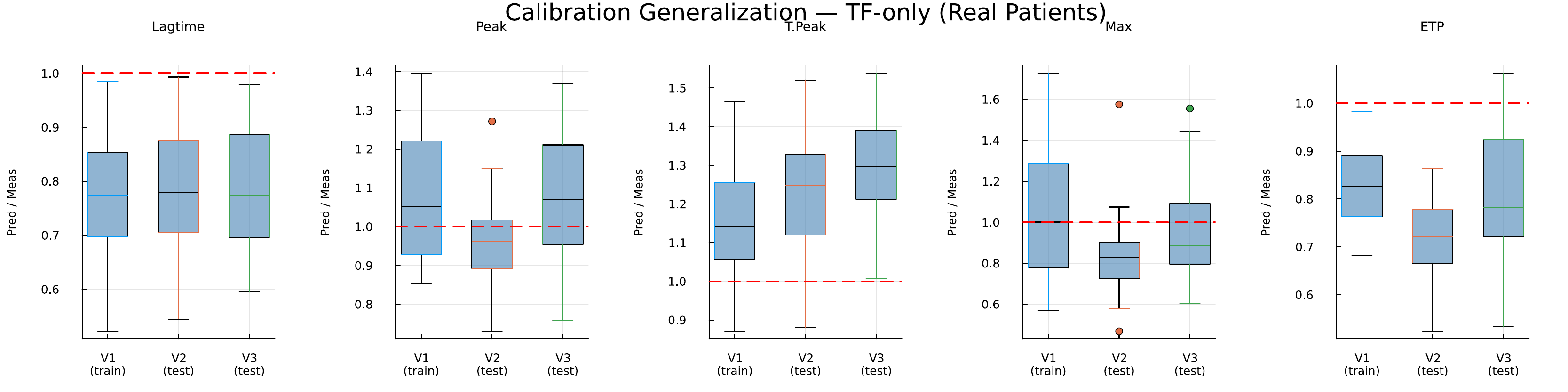}\\[6pt]
\includegraphics[width=\textwidth]{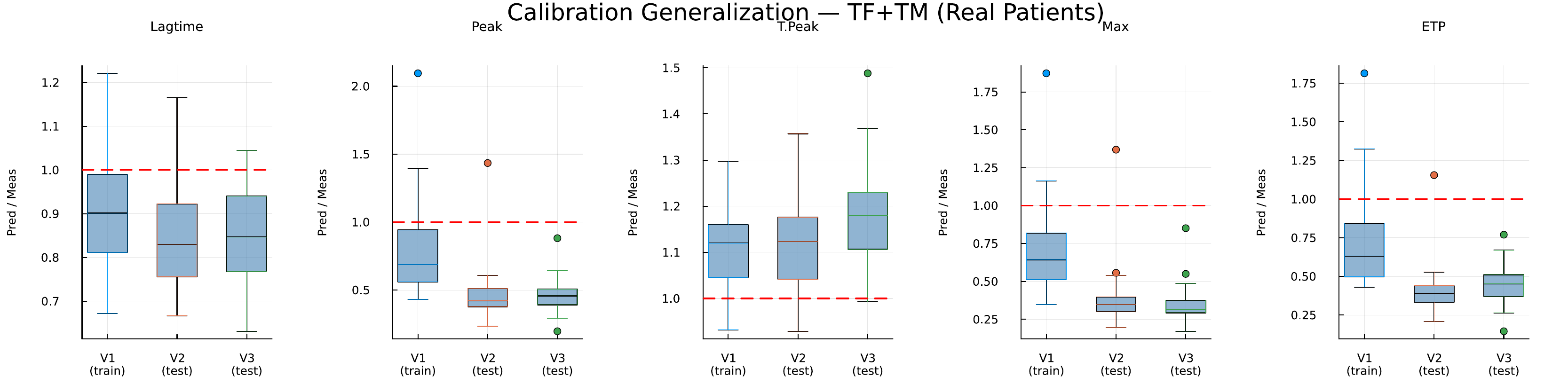}
\caption{\textbf{Cross-visit calibration generalization.} Boxplots of the
ODE-predicted/dataset ratio by visit for real patients. The model was calibrated
on Visit~1 only. Top: under TF-only conditions, ratios remained near 1.0 at Visits~2
and~3. Bottom: under TF+TM conditions, the ratios differed more across visits,
particularly for peak and maximum rate.}
\label{sfig:generalization}
\end{figure}

\begin{figure}[ht]
\centering
\includegraphics[width=0.48\textwidth]{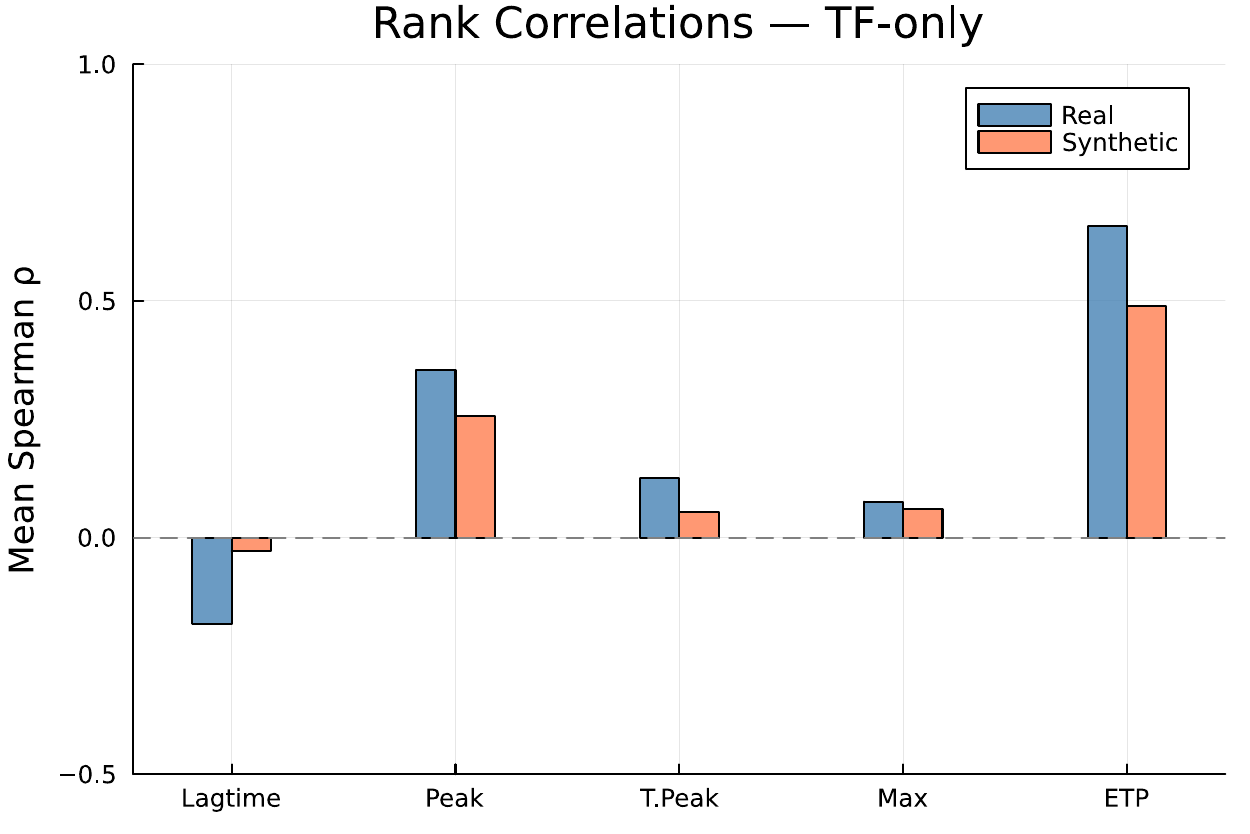}
\hfill
\includegraphics[width=0.48\textwidth]{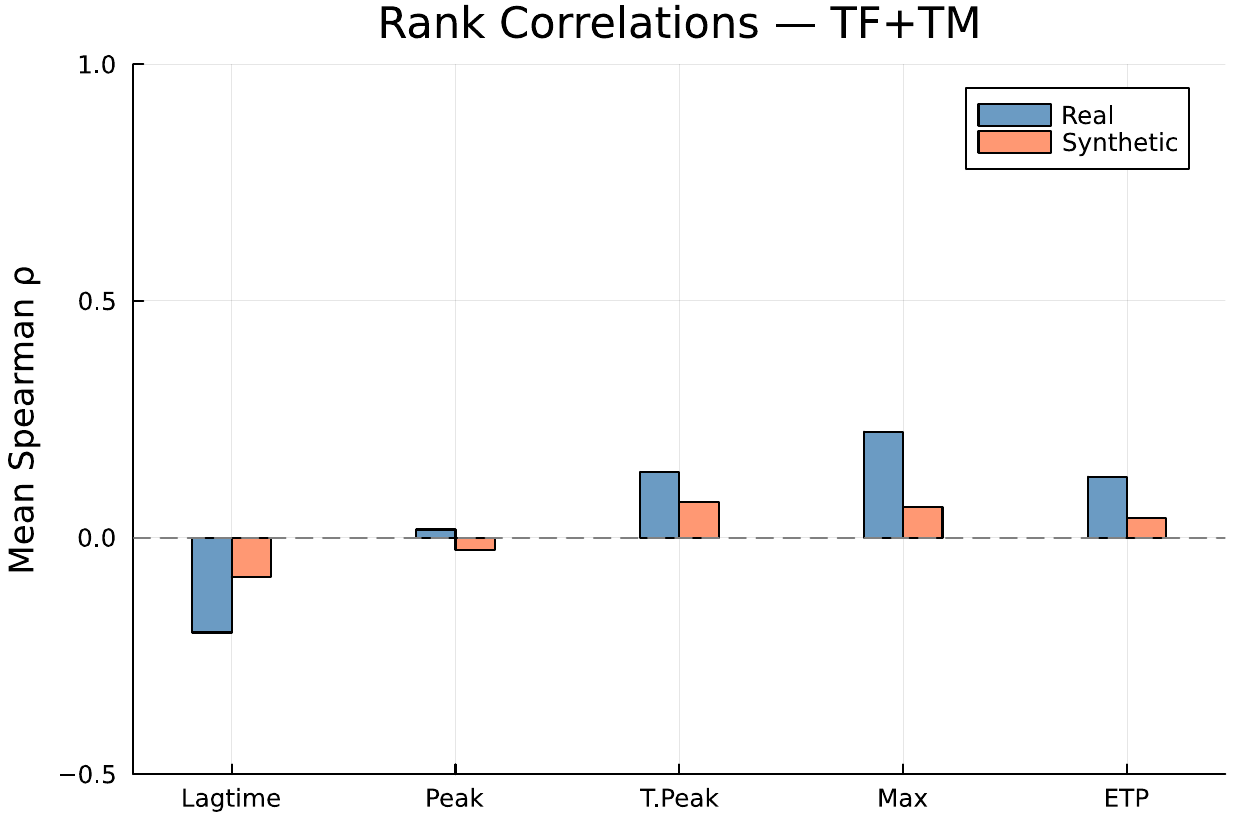}
\caption{\textbf{Spearman rank correlations between ODE-predicted and dataset
TGA features.} Left: TF-only. Right: TF+TM. ETP is the best-predicted feature
($\rho \approx 0.6$--$0.8$ for real patients under TF-only). Weak rank
correlations for other features are expected given the 5-parameter
population-level calibration.}
\label{sfig:rankcorr}
\end{figure}

\begin{figure}[ht]
\centering
\includegraphics[width=\textwidth]{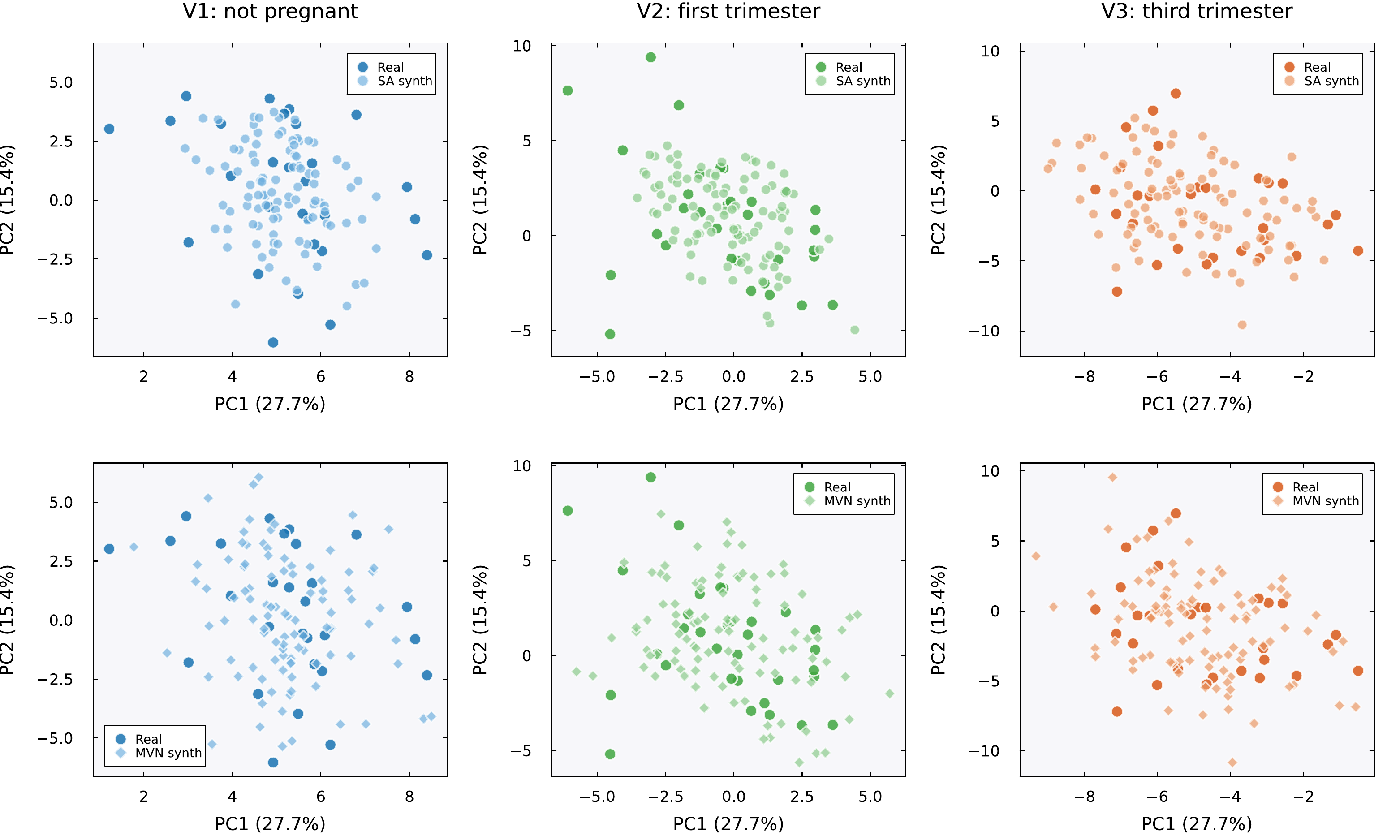}
\caption{\textbf{PCA projections by visit: SA and MVN.} Each panel shows the
first two principal components (PC1: 27.7\% variance, PC2: 15.4\%) computed
from standardized per-visit real patient data ($K{=}23$ patients per visit,
72 features). Dark markers indicate real patients; lighter markers indicate
synthetic patients ($N{=}100$). Within each visit (column), the SA panel (top)
and MVN panel (bottom) share identical axis ranges, so the relative dispersion
of the two methods can be read against the same real data cloud. Top row:
SA-generated patients (circles)
cluster tightly around the real data cloud at all three visits. Bottom row:
MVN-generated patients (diamonds) show greater dispersion, particularly at
Visits~2 (first trimester) and~3 (third trimester), where MVN patients extend
beyond the convex hull of the real data. This excess scatter reflects the
spurious variance introduced by Ledoit--Wolf regularization of the
rank-deficient covariance (Fig.~\ref{sfig:eigenvalue-spectrum}). Colors encode visit: blue
(V1, not pregnant), green (V2, first trimester), orange (V3, third trimester).}
\label{sfig:pca-by-visit}
\end{figure}

\FloatBarrier

\subsection*{Sampling correctness and privacy diagnostics}
\label{ssec:sampling-correctness}

\begin{figure}[ht]
\centering
\includegraphics[width=\textwidth]{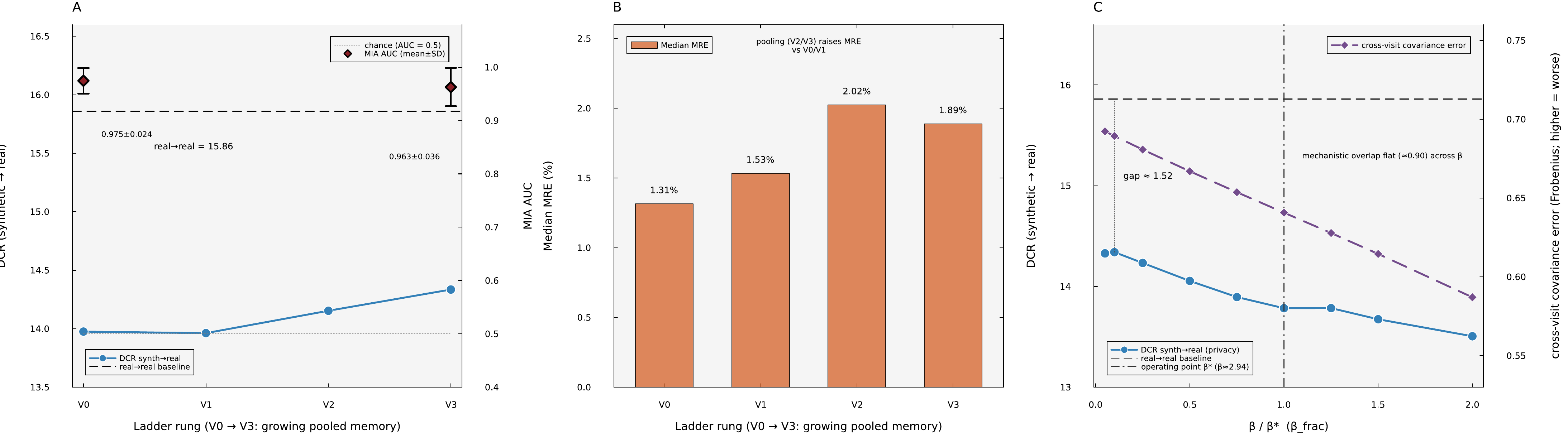}
\caption{\textbf{Sampling correctness and the membership signal.}
(A) Across V0--V3, DCR rose slightly but stayed below the real-to-real baseline, while
membership-inference AUC remained $0.96$--$0.97$. (B) Median relative error across the four
procedures. (C) Across a $40$-fold inverse-temperature sweep around $\beta^*$, DCR increased
as $\beta$ fell but remained below the real-to-real baseline. Cross-visit covariance error
increased, while mechanistic overlap changed little.}
\label{sfig:sampling-ladder}
\end{figure}

We asked whether a synthetic cohort could reveal that an already-known de-identified
assay profile was stored in the memory matrix. The source data had no direct identifiers or
links to named individuals, so this was not a test of personal identification. In standardized
$216$-dimensional space, the median distance to the closest real record (DCR) was $13.8$ for
synthetic profiles, compared with a leave-one-out real-to-real distance of $15.9$. We then
tested $40$ random splits with $15$ stored and $8$ held-out profiles. Distance to the synthetic
cohort distinguished the groups with mean AUC $0.97$ (standard deviation $0.02$; $0.5$ meant
no discrimination and $1.0$ perfect discrimination). To test whether the numerical sampler
caused this signal, we compared the reported sampler with sphere initialization, pooled samples
after burn-in and thinning, and Metropolis-adjusted sampling
(Table~\ref{stab:sampling-ladder}, Fig.~\ref{sfig:sampling-ladder}). Metropolis acceptance was
$0.999$, and the AUC remained $0.96$--$0.97$. Median DCR changed only from $14.0$ to $14.3$ and
remained below $15.9$. Changing the Markov-chain procedure therefore did not remove the
membership signal.

We also varied the inverse temperature and sampled the target distribution directly.
Across a $40$-fold sweep around $\beta^*$, lower $\beta$ increased median DCR but it plateaued
near $14.3$; cross-visit covariance error increased from $0.59$ to $0.69$
(Fig.~\ref{sfig:sampling-ladder}C). The closed-form stochastic-attention target in
Eq.~\maineqref{eq:exact_mixture} is the exact mixture
$\sum_k q_k\mathcal{N}(\mathbf{m}_k,\beta^{-1}\mathbf{I})$, with
$q_k=r_k/\sum_j r_j$ for unit-normalized memories. Its direct sampler requires only a component
draw and isotropic Gaussian noise. It has no initialization, burn-in, or discretization error.
A matched direct-sampler cohort gave median relative error $1.68\%$, cross-visit error $0.60$,
mechanistic overlap $0.91$, median novelty $0.50$, and median DCR $13.87$, all within the range
of the four Markov-chain procedures. Across $100$ direct-sampler cohorts, median relative error was
$1.40\%$ (5th--95th percentile $0.97$--$1.91\%$) and median DCR was $13.95$
($13.61$--$14.39$). Exact sampling gave AUC $0.98$ over the same $40$ splits. This showed that
the membership signal came from the target distribution, not from the numerical sampler. At
this cohort size, the test could therefore distinguish whether an already-known de-identified
profile was stored regardless of how the target distribution was sampled.

\begin{table}[ht]
\centering
\small
\caption{\textbf{Sampler comparison.} All procedures used the same $\beta^*$, memory,
and decoder. We report median relative error (MRE), cross-visit correlation error,
mechanistic overlap, novelty, DCR, and membership-inference AUC where evaluated. The
real-to-real DCR baseline was $15.9$. The AUC used the same $40$ splits for V0, V3, and the
direct stochastic-attention sampler. \emph{Exact} drew directly from
Eq.~\maineqref{eq:exact_mixture} and was not a
Markov-chain procedure. Across $100$ direct-sampler cohorts, median MRE was $1.40\%$ (5th--95th
percentile $0.97$--$1.91\%$) and median DCR was $13.95$ ($13.61$--$14.39$).}
\label{stab:sampling-ladder}
{\small
\setlength{\tabcolsep}{4pt}
\begin{tabular}{llcccccc}
\hline
Rung & Sampler & MRE (\%) & Cross-visit Frob. & Mech. overlap & Novelty & DCR & MIA AUC \\
\hline
V0 & anchor, endpoint & 1.31 & 0.65 & 0.89 & 0.51 & 13.97 & 0.97 \\
V1 & sphere init      & 1.53 & 0.57 & 0.89 & 0.50 & 13.96 & n/a  \\
V2 & pooled ULA       & 2.02 & 0.59 & 0.88 & 0.50 & 14.15 & n/a  \\
V3 & pooled MALA      & 1.89 & 0.62 & 0.92 & 0.52 & 14.33 & 0.96 \\
\hline
Exact & analytic ancestral & 1.68 & 0.60 & 0.91 & 0.50 & 13.87 & 0.98 \\
\hline
\end{tabular}
}
\end{table}

\begin{figure}[ht]
\centering
\includegraphics[width=\textwidth]{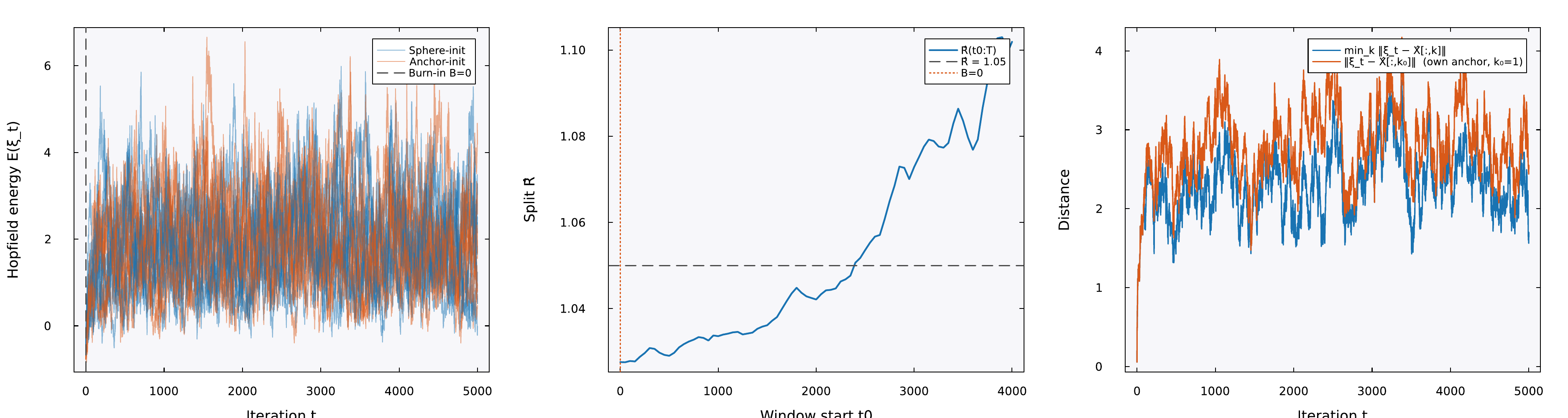}
\caption{\textbf{Mixing diagnostic at $\beta^*$.} Energy traces, the between-chain
$\hat{R}$ statistic, and the anchor-distance trajectory for sphere-initialized Langevin
chains. The target distribution at $\beta^*$ was multimodal, with one component centered on
each stored pattern. Metropolis acceptance near $0.999$ indicated negligible discretization
bias.}
\label{sfig:mixing-diagnostic}
\end{figure}

\end{document}